\documentclass[lettersize,journal]{IEEEtran}
\usepackage{amsmath,amsfonts}
\usepackage{algorithmic}
\usepackage{algorithm}
\usepackage{array}
\usepackage[caption=false,font=normalsize,labelfont=sf,textfont=sf]{subfig}
\usepackage{textcomp}
\usepackage{stfloats}
\usepackage{url}
\usepackage{verbatim}
\usepackage{graphicx}
\usepackage{tabularx}
\usepackage{cite}
\usepackage{float}
\usepackage{multirow}
\usepackage{diagbox}
\usepackage{amssymb}
\usepackage{color,xcolor}
\usepackage{booktabs}
\usepackage{rotating}
\usepackage[table]{xcolor}
\usepackage{soul}
\usepackage{xcolor}
\sethlcolor{yellow}
\usepackage[utf8]{inputenc}
\usepackage[font=small,skip=2pt]{caption}
\usepackage[utf8]{inputenc}
\soulregister\cite7
\soulregister\citet7
\soulregister\citep7
\soulregister\ref7

\usepackage[font=small,skip=4pt]{caption}           % skip=<标题与图形/表格主体之间的垂直距离>
\usepackage[most]{tcolorbox} 
\newtcolorbox{highlightboxyellow}{
  colback=yellow!70,          % background define as 30% light yellow
  colframe=yellow!80!black,    % margin color setup
  boxrule=0pt,                % set margin as 0 pt
  breakable,                  % allow cross the page
  left=0pt, right=0pt, top=0pt, bottom=0pt
}
\newtcolorbox{highlightboxblue}{
  colback=blue!20,          
  colframe=blue!50!black,   
  boxrule=0pt,              
  breakable,                
  left=0pt, right=0pt, top=0pt, bottom=0pt
}

\newtcolorbox{highlightboxgreen}{
  colback=green!20,          
  colframe=green!50!black,   
  boxrule=0pt,               
  breakable,                 
  left=0pt, right=0pt, top=0pt, bottom=0pt
}

\newcommand{\hlnum}[1]{%
  \begingroup
    \setlength\fboxsep{1pt}%
    \rlap{\colorbox{gray!20}{\strut\hphantom{#1}}}%
    #1%
  \endgroup
}

\begin{document} 

\title{Deep-Learning-Based Control of a Decoupled Two-Segment Continuum Robot for Endoscopic Submucosal Dissection

\author{ Yuancheng Shao$^{*}$$^{1,2}$, Yao Zhang$^{*}$$^{3}$, Jia Gu$^{2}$, \emph{Senior Member, IEEE}, Zixi Chen$^{4}$, Di Wu$^{5}$, Yuqiao Chen$^{2,6}$, Bo Lu$^{7}$, \emph{Member, IEEE}, Wenjie Liu$^{1}$, Cesare Stefanini$^{4}$, \emph{Member, IEEE},  Peng Qi$^{1}$, \emph{Member, IEEE}  % <-this % stops a space
\thanks{This work is supported by the National Key Research and Development Program of China under Grant No. 2023YFB4705200, in part by the National Natural Science Foundation of China under Grant No. 62273257 and 62203315, in part by the Natural Science Foundation of Jiangsu Province of China (BK20220490), in part by the Innovation and Entrepreneurship Leading Talent Program of Suzhou City (ZXL2023156), and in part by the Young Elite Scientists Sponsorship Program by CAST (2023QNRC001). \emph{(Yuancheng Shao and Yao Zhang contributed equally to this work.)} \emph{(Corresponding Author: Peng Qi and Bo Lu)}.}
\thanks{$^{1}$ Department of Control Science and Engineering, College of Electronics and Information Engineering, Tongji University, No. 1239, Siping Road, Shanghai 200092, China;} 
\thanks{$^{2}$  Department of Data Science, City University of Macau, China;
}
\thanks{$^{3}$ Department of Mechanical Engineering, KU Leuven, 3000 Leuven, Belgium;}
\thanks{$^{4}$ The Biorobotics Institute and the Department of Excellence in Robotics and AI, Scuola Superiore Sant’Anna, 56127 Pisa, Italy;}
\thanks{$^{5}$ The Maersk Mc-Kinney Møller Institute, University of Southern Denmark, 5230 Odense, Denmark;}
\thanks{$^{6}$ Surgical Robotics R\&D Center, Zhuhai Institute of Advanced Technology, Zhuhai 519085, Guangdong Province, China;}
\thanks{$^{7}$ The Robotics and Microsystems Center, School of Mechanical and Electric Engineering, Soochow University, Suzhou, Jiangsu, China.}
} 

}

\markboth{ASME/IEEE Transactions on Mechatronics,~Vol.~x, No.~x, ~2024}%
{Shell \MakeLowercase{\textit{et al.}}: A Sample Article Using IEEEtran.cls for IEEE Journals}


\maketitle

\begin{abstract}
% \fcolorbox{gray}{yellow}
Manual endoscopic submucosal dissection (ESD) is technically demanding, and existing single-segment robotic tools offer limited dexterity. These limitations motivate the development of more advanced solutions. To address this, DESectBot, a novel dual-segment continuum robot with a decoupled structure and integrated surgical forceps, enabling 6 degrees of freedom (DoFs) tip dexterity for improved lesion targeting in ESD, was developed in this work. Deep learning controllers based on gated recurrent units (GRUs) for simultaneous tip position and orientation control, effectively handling the nonlinear coupling between continuum segments, were proposed. The GRU controller was benchmarked against Jacobian-based inverse kinematics, model predictive control (MPC), a feedforward neural network (FNN), and a long short-term memory (LSTM) network. In addition, a systematic ablation study under identical training conditions showed that, for the considered control task, the GRU achieves better generalization, faster convergence, and higher computational efficiency than the LSTM. In nested-rectangle and Lissajous trajectory tracking tasks, the GRU achieved the lowest position/orientation RMSEs: 1.11 mm/ 4.62° and 0.81 mm/ 2.59°, respectively. For orientation control at a fixed position (four target poses), the GRU attained a mean RMSE of 0.14 mm and 0.72°, outperforming all alternatives. In a peg transfer task, the GRU achieved a 100\% success rate (120 success/120 attempts) with an average transfer time of 11.8 s, the STD significantly outperforms novice-controlled systems. Additionally, an \textit{ex vivo} ESD demonstration—grasping, elevating, and resecting tissue as the scalpel completed the cut—confirmed that DESectBot provides sufficient stiffness to divide thick gastric mucosa and an operative workspace adequate for large lesions.These results confirm that GRU-based control significantly enhances precision, reliability, and usability in ESD surgical training scenarios.

% Endoscopic Submucosal Dissection (ESD) has become a widely embraced clinical method for the removal of precancerous and cancerous lesions for the treatment of early-stage gastrointestinal (GI) cancer. In this paper, we designed and manufactured a decoupled two-segment continuum robotic system named DESectBot for ESD and investigated a deep learning based control method called Gate Recurrent Gate (GRU) on DESectBot for challenging tasks such as precise simultaneous position and orientation control. To compare with GRU, two model-based Jacobian Control and Model Predictive Control (MPC) methods were implemented as well. The results validated that the GRU controller outperforms the other two model-based controllers in the trajectory tracking test and has a performance of at least 15.74\% higher than that of the MPC and far exceeds that of the ordinary Jacobian controller in orientation control. Considering that, a further phantom study combined with two identical DESectBots based on GRU controller were implemented, which successfully performed real ESD operations in only 23 seconds, such as grasping, stripping, and lifting. In summary, given its precise accuracy for simultaneous control of position and orientation and the further success of the phantom study, DESectBot with the proposed GRU controller could be a potential combination for ESD treatment.

\end{abstract}

\begin{IEEEkeywords}
Endoscopic Submucosal Dissection (ESD), continuum robot, learning-based control
\end{IEEEkeywords}
\vspace{-0.15in}
\section{Introduction}

%可以先介绍两句Gastrointestinal (GI) cancer ，用一些数据证明该疾病有危险，患病人数多等特征，类似于我如下这段话（选自其他论文）
%Gastrointestinal (GI) cancer is gradually becoming high incidence worldwide, including esophageal cancer, gastric cancer, colorectal cancer, etc. The high incidence and low diagnosis rate of this type of cancer have serious implications for human health. According to the “Global Cancer Report” released by the World Health Organization in 2018 [1], gastric cancer, one of the gastrointestinal cancers, accounts for 5.7% of all cancer cases and ranks fifth. Due to the low diagnosis rate of gastrointestinal cancer, people often lose the best time for timely treatment because they discover the cancer too late. According to clinical practice, the depth of invasion of early malignant tumors is limited to the mucosal layer or exceeds the submucosal layer. If it can be diagnosis and treatment early, it can effectively improve the five-year survival rate of patients, reaching a level of 90%. However, for advanced gastric cancer patients, this data is only 10% [2]. 
% \fcolorbox{gray}{yellow}{R3.11}
Gastrointestinal (GI) cancers represent a major global health burden. In 2018, an estimated 4.8 million new cases led to 3.4 million deaths, accounting for 26 \% of all cancer diagnoses and 35 \% of cancer fatalities worldwide\cite{arnold2020global}. These high mortality rates are driven by late‐stage detection and a range of risk factors—including infections, lifestyle behaviors, and dietary habits \cite{hormati2022gastrointestinal}.These alarming mortality figures underscore the urgent need for progress in prevention, early diagnosis, and the development of more effective treatments to address this high-mortality challenge to public health.

% Gastrointestinal (GI) cancer poses a significant health threat globally. In 2018, approximately 4.8 million new cases of \fcolorbox{gray}{yellow}{R3.20}\hl{GI} cancers  were reported worldwide, resulting in 3.4 million related deaths, with GI cancers representing 26\% of cancer incidence and 35\% of cancer-related mortality worldwide . Ranked as leading causes of cancer-related mortality, their impact is exacerbated by late detection and various risk factors, including infections, lifestyle choices, and dietary patterns . 

Endoscopic Submucosal Dissection (ESD) has become a widely embraced clinical method for treating early-stage GI cancer, offering a significant boost to survival rates \cite{oka2006advantage}. This minimally invasive surgical procedure not only minimizes patient discomfort but also facilitates faster recovery times and shorter hospital stays. In contrast to conventional endoscopic mucosal resection (EMR), ESD offers several advantages: (\romannumeral1) lower recurrence rates; (\romannumeral2) the ability to excise lesions en bloc regardless of size, whereas EMR is confined to lesions under $20$ mm; (\romannumeral3) provision of optimal specimens for precise histopathological evaluation; and (\romannumeral4) potential resection of lesions unsuitable for EMR \cite{kotzev2019master}. However, performing ESD requires advanced technical skills and involves a higher risk of complications, necessitating years of training to master the various modalities of procedure \cite{teoh2010difficulties}.
\vspace{-0.2in}
\subsection{Robot Assisted Endoscopy}

However, traditional endoscopes face several limitations. Firstly, their limited maneuverability and precision can make it difficult to accurately target and resect early-stage cancers or precancerous lesions, particularly in the anatomically complex regions of the GI tract \cite{fujishiro2008perspective}. Secondly, the success of these procedures is contingent upon the clinician’s manual dexterity and critical decision-making, which contributes to variability in ESD outcomes\cite{zhang2020learning}. Lastly, most endoscopes are still manually controlled, which places a significant physical strain on physicians during lengthy ESD procedures. This can result in fatigue, decreased precision, and potentially compromise the treatment's success. Therefore, to address the aforementioned issues, or at least some of them, researchers have introduced robotics technology into the endoscopy procedure.

%In \cite{shang2011articulated}, the $iSnake$ robot, embedded with micromotors and local tendons in each joint, was designed for single-port surgery. Building on this, an enhanced version $i^{2}Snake$ aimed at improving manipulation force at the robot's tip was introduced \cite{berthet20182}. For the treatment of the \fcolorbox{gray}{yellow}{R3}\hl{GI} region, \cite{hwang2020k} presented a flexible surgical robot platform named K-FLEX. To ease the procedure of the ESD treatment, a master-slave dual-arm robotics platform was proposed in \cite{lau2016flexible}.% 
To mitigate potential patient harm and access lesions deep within the body through natural orifices, research on soft and flexible continuum surgery instruments has gained significant traction \cite{gu2023survey}.  \cite{gao2024transendoscopic} introduced a transendoscopic flexible parallel continuum robotic mechanism with a miniature wrist for ESD treatment. However, single-segment robotic endoscopes have limited flexibility and range of motion, and they also struggle to maintain stable visualization. This becomes particularly problematic in complex areas of the GI tract, such as the esophago-gastric junction. This region, where the esophagus meets the stomach, features a sharp angle and tight space, making it challenging to access and resect lesions with a single-segment endoscope due to its limited maneuverability \cite{rice2017esophagus}. Achieving and maintaining a direct line of sight for precise dissection can be difficult. Furthermore, manipulating instruments for cutting and coagulation without compromising the visual field or risking injury to surrounding tissues poses a significant challenge, increasing the likelihood of incomplete resection or perforation.

To address the aforementioned challenges, the development of multi-segment robotic endoscopes has been investigated. Compared to the single-segment robot, the multi-segment robot has advantages with more DoFs and higher distal dexterity \cite{burgner2015continuum}. %To facilitate traversing the complex and narrow bronchial lumen of the lung, a three-segment robot was designed for pulmonary intervention as described in \cite{kato2020robotized}. Similarly, to enhance endoscope maneuverability, featuring module bending segments was introduced for colonoscopy in \cite{chen2014modular}. 
In the context of ESD treatment, multi-segment robotic endoscopes show promising capabilities for precise execution of surgical steps, including multi-angle cutting operations relative to lesion surfaces and efficient en-bloc resection. Notably, a two-segment robotized soft endoscope incorporated with two calibrated cameras demonstrated feasibility for ESD treatment \cite{chen2023robotized}. Moreover, our previous work introduced a dual-segment robotic platform, DESectBot, featuring a spatial cross-curved disk skeleton structure and showed promising trajectory tracking performance. However, the multi-segment structure introduces increased nonlinearities, and the need for more actuators to control each segment adds complexity to control. 

\vspace{-0.2in}
\subsection{Deep Learning Based (DL-based) Control of Continuum Robot}
Deep learning (DL) methods are highly valued in the control of continuum robots due to their ability to model complex and non-linear dynamics and use data to improve precision and accuracy. DL-based algorithms could continuously refine control strategies for improved performance, particularly in applications requiring intricate manipulation, such as minimally invasive procedures. To precisely control tendon-driven robots, the Recurrent Neural Network (RNN) method is proposed to address the non-linear and non-repeatable elastic behavior \cite{choi2020hybrid}. Furthermore, two DL methods, Multilayer Perceptron (MLP) and RNN, were utilized to predict the distal-end force of tendon-sheath mechanisms in flexible endoscopic surgical robots \cite{li2019distal}. To model hysteresis and other nonlinearities in flexible continuum robots, both Long Short‑Term Memory (LSTM) and Gated Recurrent Unit (GRU) networks—two memory‑based recurrent architectures—have been explored. LSTM was first used to predict distal‑tip responses and, in a subsequent study, embedded in a feed‑forward controller to compensate for hysteretic behavior in free and constrained motions \cite{wu2021hysteresis,wu2022deep}. More recently, GRU networks have likewise been implemented to control multi‑DOF articulated soft robots, demonstrating comparable efficacy for handling nonlinear, hysteretic dynamics \cite{schafke2024learning, feng2021learning}. Furthermore, a combination of DL methods and an online optimization kinematics controller was proposed to further reduce the errors caused solely by the neural network controller \cite{chen2023hybrid}. However, as of now, DL methods have yet to be utilized in the control of two-segment endoscope robots. Consequently, this paper seeks to leverage DL techniques to address the nonlinear phenomena in the DESectBot system and further achieve precise control of the robot, showcasing the potential for advanced DL techniques to enhance robotic endoscopy.
\vspace{-0.2in}

\subsection{Main Contributions and Paper Structure}

% To the best of the authors' knowledge, this is the first time that the control of a two-segment endoscopic robot is implemented using a DL method. 

To the best of our knowledge, this is the first paper to leverage a recurrent-based deep learning method on a two-segment endoscopic robot for challenging tasks such as simultaneous position and orientation control, which is required by surgery such as endoscopic submucosal dissection. 

The main contributions of this paper are:
\begin{itemize}
\item A novel dual‐segment continuum robot DESectBot with an integrated surgical forceps end effector was established for ESD. %%最后再润色和改
\item  Implement a proposed data-driven method to model the inverse kinematics of DESectBot, from which a DL-based Gate Recurrent Unit (GRU) controller has been established. Validation of the position and orientation control precision of DESectBot integrated with GRU, by comparing several model-based and data-driven methods, through benchtop experiments.
%%% Add another validation to modify this 
\item Carry out the benchmark peg transfer test by GRU-driven DESEctBot and make a performance comparison with a well-known surgical robot which confirms its suitability for standardized ESD skill training.
\end{itemize}
The structure of this paper is outlined as follows: Section II details DESectBot’s hardware and kinematics. Section III presents the GRU‑based controller and data collection, contrasting model‑based and learning approaches. Section IV reports trajectory‑tracking and dexterity tests. Section V evaluates a benchmark peg‑transfer task and \textit{ex vivo} ESD demonstration. Section VI concludes and outlines future work.

\section{System design}

\begin{figure*}
  \centering
\includegraphics[width=0.95\textwidth]{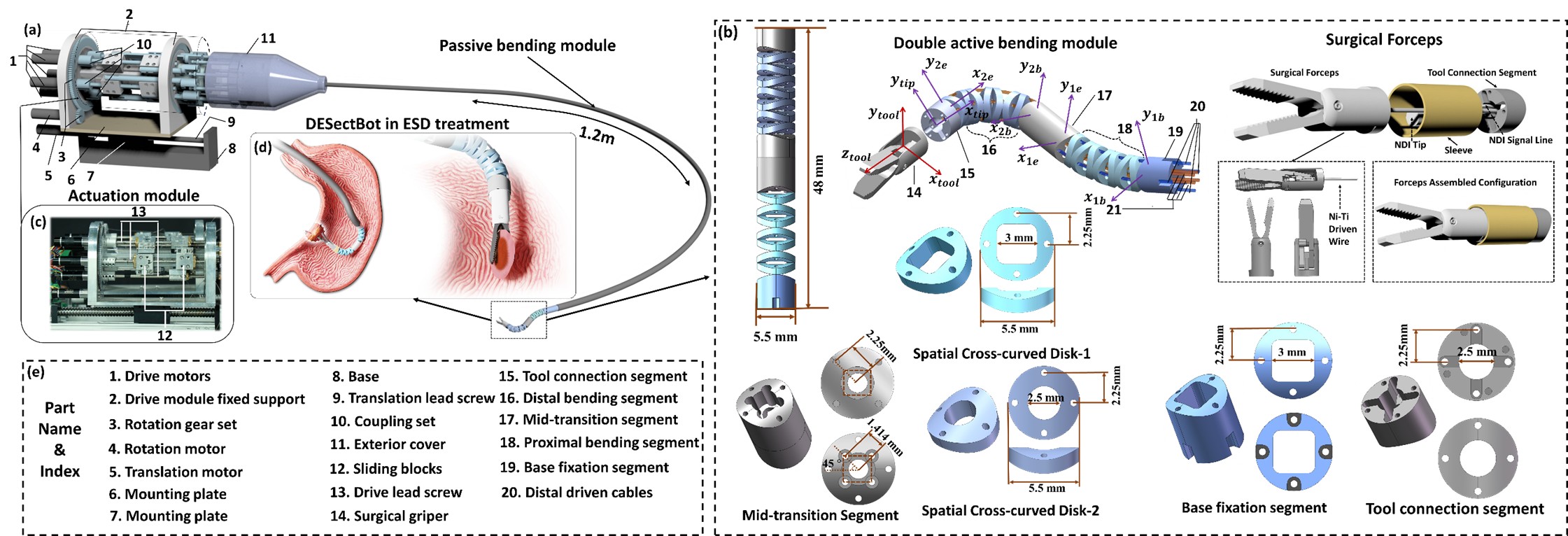}
  \caption{Decoupled Endoscopic Submucosal Dissection Robot (DESectBot): (a) detailed assembly drawing of DESectBot system, (b) detailed parts drawing of double active bending module with surgical gripper in kinematics coordinate system, (c) DESectBot actuation module, (d) demonstration of DESectBot in ESD treatment, (e) part name and corresponding index}
\label{desectbot}
\vspace{-0.2in}
\end{figure*}

% The overall DESectBot system is composed of an actuation module, an exterior cover, a passive bending module and a double active bending module, with 6 DOFs to satisfy the requirements of ESD. The DESectBot basic frame is made of light stainless steel. The control system contains seven DC motors (ECXSP16L, Maxon Motor AG, Switzerland) with the specification gearbox GPX16HP (44:1) and is controlled through a multi-axis motion controller (P-MASON0N, Elmo Motion Control Ltd, Israel).  The exterior cover is 3D printed and can be easily disassembled. 

% \begin{highlightboxyellow}
Commercial multifunctional tools such as marking needles for submucosal injection, mucosal incision knives, electrosurgical dissection knives and hemostatic devices are widely used \cite{maple2015endoscopic}, yet they generally lack full six DoFs capability. This shortfall hinders precise lesion targeting and accurate tool‐tip orientation while maintaining endoscope stability within the gastrointestinal lumen \cite{kohli2019endoscopes}. As a result, clinicians must undergo extensive training to execute these complex maneuvers safely and effectively \cite{ma2018endoscopic}, and inadequate control can lead to tissue injury, perforation or bleeding \cite{misumi2021prevention}. It is therefore essential to develop ESD instruments with true 6 DoFs dexterity in order to  to meet clinical demands for lesion access and shorten the learning curve, reduce procedure time and enhance surgical safety \cite{simaan2018medical}.

% \end{highlightboxyellow}
The overall DESectBot system addresses these requirements by providing 6 DoFs via four integrated modules: an actuation module (encompassing translation and rotation) enclosed by an exterior cover, a passive bending module, a double active bending module (DABM) and a custom-designed end effector.

\subsubsection{DESectBot actuation module}
%% 
% The actuation module is actuated by seven DC motors (ECXSP16L, Maxon Motor AG, Switzerland) with the specification gearbox GPX16HP (44:1) and is controlled through a multi-axis motion controller (P-MASON0N, Elmo Motion Control Ltd, Israel).
% \fbox{1}

The actuation module consists of seven DC motors (ECXSP16L, Maxon Motor AG, Switzerland) equipped with GPX16HP (44:1) gearboxes. These motors are divided into three groups: a drive motor unit, a rotation motor, and a translation motor. All motors are controlled by a multi-axis motion controller (P-MASON0N, Elmo Motion Control Ltd, Israel). The drive motor unit includes five DC motors, each connected to a coupling and cable fixing slider blocks. Using a lead screw and nut mechanism, the rotational motion of the DC motors is converted into the linear motion of the sliders. These sliders are arranged in a regular pentagon along the central axis. Four sliders actuate the double active bending module via nickel-titanium (NiTi) tendons ($\phi$ $0.5$ mm), while the remaining slider drives the surgical tool. The entire actuation module can rotate and translate via the rotation and translation motors, respectively. The module is enclosed in a 3D-printed exterior cover, which is easily disassembled for maintenance.

% The actuation module has five sets of couplings and cable fixing sliders. The sliders are set uniformly along the central axis in a regular pentagon. Four sets of them are used to drive the double active bending module, and the remaining set is employed to control the driven cable of the surgery tool. The nickel-titanium (NiTi) tendons with a diameter of 0.5 mm are used to actuate the active bending module, with the cable fixation slider at the same distance from the arm center axis. This configuration ensures the manipulator arm itself has four DOFs and can achieve various motions with two segments. Furthermore, the rotation motion module and linear motion module can drive the manipulator arm to rotate around its axis and move forward and backward. Overall, with the cooperation of all the above-driven units, the DESectBot has 6 DOFs and can achieve end position and orientation simultaneously.

\subsubsection{DESectBot passive bending module}
The passive bending module is made of ebonite and has a standard length of $1.2$ meters. NiTi tendons are threaded through this sheath structure.

% The sheath structures that drive the double active bending module pass through the passive bending module.
%%%%%%%%%%%%%%%%%%%%%%%%%%%%%%%%%%%%%%%%%%%%%%%%%%%%%%%%%%%%%%%%% Merge 3) and 4）content and illustrate decoupling
\subsubsection{DESectBot DABM}
As shown in Fig. \ref{desectbot} (b), the DABM consists of a proximal and a distal bending segment connected by a rigid mid-transition segment. Each segment is actuated by NiTi tendons, allowing it to bend independently in isolation. The mid-transition link mechanically decouples the two segments, so bending one segment minimally affects the other. Both bending segments utilize a dual continuum manipulator mechanism based on spatial cross-curved disk (SCD) joints: a skeletal element permitting continuous multi-directional bending. The proximal segment uses an SCD-1 joint skeleton and the distal segment uses SCD-2. Each SCD disk has two perpendicular curved slots, forming a cross-curved structure that permits multi-directional bending. In the proximal segment, tendon guide holes are evenly spaced (every 90°), whereas in the distal segment they are rotated by 45°. With two pairs of opposing tendons per segment, this SCD-based structure enables full 3D bending of both segments. Stacking the disks in alternating orientations (each pair rotated 90° relative to the next) yields a cross-arc configuration that allows the DABM to bend in four orthogonal directions with minimal inter-segment coupling.

\subsubsection{Surgical Forceps End Effector}
In this study, as shown in Fig. \ref{desectbot}(b) a surgical forceps with 5mm diameter and 20mm length was designed and integrated as the DESectBot’s end effector at the distal tip of the DABM. The forceps attaches via a custom 6.5mm-diameter, 15mm-long sleeve and incorporates a compact antagonistic NiTi cable-driven mechanism for precise tissue manipulation. One 6 DoFs electromagnetic (EM) tracker (Aurora V3.1, Northern Digital Inc., Canada) tip is embedded in the tool segment to provide real-time tip pose feedback. The surgical forceps assembly is modular, allowing quick replacement and sterilization between procedures. When coupled with DABM, the integrated forceps achieve enhanced dexterity and control, enabling simultaneous yet decoupled control of position and orientation. The decoupling mechanism between the proximal and distal segments is detailed in previous work \cite{liu2024desectbot}, where it was shown that the unintended motion (crosstalk) between the segments is minimal, with angular deviations of less than 1°, confirming effective decoupling. This capability overcomes the limited dexterity and orientation control of conventional ESD instruments, simplifying ESD procedures and improving lesion-targeting precision and tissue manipulation.
% By leveraging the DABM flexibility and advanced control capabilities, this integrated design overcomes the limited dexterity and orientation control of conventional ESD instruments. Consequently, the DESectBot with integrated forceps simplifies ESD procedures, improves lesion-targeting precision and tissue manipulation.

% \subsubsection{Hysteresis Analysis}
% \fcolorbox{gray}{yellow}{R2.6, R2.7}\hl{Because DESectBot relies on multiple NiTi tendons routed through a long passive bending module, the system is inherently prone to hysteresis, frictional losses, and loading–unloading drift commonly observed in tendon–sheath mechanisms. To quantify these effects, we conducted controlled bidirectional actuation tests on both bending segments under varying amplitudes and repeated cycles, using the 6-DoF NDI tracker to record ground-truth tip motion. The analysis will include: (i) tendon command–tip pose hysteresis loops to extract loop area as a measure of nonlinear lag; (ii) loading vs. unloading path deviation to distinguish friction-dominated from elasticity-dominated components; (iii) multi-cycle drift evaluation to assess accumulated pose deviation.}

\subsubsection{Hysteresis Analysis}
% \fcolorbox{gray}{yellow}{R2.6, R2.7}\hl{Because DESectBot relies on multiple NiTi tendons routed through a long passive bending module, hysteresis, friction, and load-dependent drift are inherent to the tendon--sheath mechanism. These effects were analyzed through controlled cyclic actuation experiments under different external loading conditions and shown in Table \ref{hysteresis analysis}. The results show predictable and non-divergent behavior under repeated loading and unloading, indicating that hysteresis effects remain bounded and sufficiently constrained to support stable and repeatable ESD-oriented manipulation.}

Because DESectBot relies on multiple NiTi tendons routed through a long passive bending module, hysteresis, friction, and load-dependent drift are inherent to the tendon–sheath mechanism. These effects were analyzed through controlled cyclic actuation experiments under different external loading conditions, confirming predictable and non-divergent behavior under repeated loading and unloading. To mitigate these effects, the system design and control framework incorporate tendon pre-tensioning, low-friction sheath routing. Regarding clinical translation, the distal tool and tendon–sheath assembly are based on biocompatible NiTi components compatible with standard sterilization methods, while motors and electronics remain outside the sterile field, supporting integration into conventional clinical workflows.
% \vspace{-0.2in}

\begin{table}[t]
\caption{Quantitative summary of drift, normalized hysteresis, and motion stroke under repeated cyclic loading}
\label{hysteresis analysis}
\centering
\setlength{\tabcolsep}{4pt}   
\renewcommand{\arraystretch}{1.05} 
\resizebox{\columnwidth}{!}{%
\begin{tabular}{c c c c}
\hline
\textbf{Load (g)} &
\textbf{Mean Drift (mm)} &
\textbf{Norm. Hysteresis Area} &
\textbf{Mean Stroke (mm)} \\
\hline
0 & 0.41 & 0.0119 & 13.40 \\
50 & 0.46 & 0.0134 & 11.16 \\
100 & 0.47 & 0.0152 & 9.72  \\
\hline
\end{tabular}}
\end{table}

\section{Robot Control Methods}
\label{methods}
In view of the fact that the DESectBot is a two-segment continuum robot system, it is a challenge to simultaneously control the position and orientation of the DABM tip to perform the ESD tasks. Therefore, two model-based control methods, three data-driven neural networks, were established here for DABM. Furthermore, the characteristics of these methods detailed analysis shown in Table \ref{tab:controller_comparison}.
\vspace{-0.1in}

\subsection{Jacobian Kinematic Control} 
The kinematic model of the DABM has already been derived under the constant-curvature assumption in our earlier work \cite{liu2024desectbot}. In that formulation, each continuum segment is parameterised by the bending‐angle vector $\Psi_i=[\theta_i,\delta_i]^T$. $\theta_i$ and $\delta_i$, and closed-form expressions for the forward and differential kinematics were obtained.

The present study builds on that foundation by incorporating the distal surgical forceps tool, which was not included in the previous model, and by developing the corresponding kinematics control strategies.  
From Fig. \ref{desectbot} (b) and Eq.~(6) in \cite{liu2024desectbot}, the representation matrix of the tool frame $\{tip\}$ relative to the base frame $\{1_b\}$ is:
\begin{equation}
  T^{1_b}_{tool} \;=\;T^{1_b}_{tip}\,T_{tool}^{tip} \;=\; T^{1b}_{1e}\,T^{1e}_{2b}\,T^{2b}_{2e}\,T^{2e}_{tip}\,T_{tool}^{tip}.
\end{equation}
Then attach a fixed forceps frame $\{tool\}$ with a tool length $d_t$ at the tip via
\begin{equation}
  T_{1_b}^{tip} \;=\;
  \begin{bmatrix}
    I_3 & p_{tool}\\[4pt]
    \mathbf{0} & 1
  \end{bmatrix},
  \quad
  p_{tool} = 
  \begin{bmatrix}0\\0\\d_t\end{bmatrix},
\end{equation}

Analogous to the continuum inverse kinematics (Eq. (5)-(7) in \cite{liu2024desectbot}), the instantaneous kinematics of the full DESectBot system is:
\begin{equation}
\dot{X}_{tool} \;=\; J_{\Psi_{tool}}\,\dot{\Psi}_{tool}
\end{equation}
Where $J_{\Psi_{tool}}$ is the Jacobian matrix of the entire continuum manipulator with forceps derived from $J_{\Psi_{i}}$.

\subsection{Model Predictive Control}

In the design of closed-loop model-based controllers for DeSectBot, Model Predictive Control (MPC) theory can be applied to cable-driven continuum robot to achieve optimal system inputs at each moment\cite{chien2021kinematic}. The proposed nonlinear model predictive control method utilizes predictions of future outputs during the modeling stage to determine the current system's optimal input. The model input is determined through a target optimization function, and this predictive model does not require multiple sensors to gather the posture of the continuum arm. It relies solely on the predicted time-series outputs to achieve precise and stable pose control of the DeSectBot.
% According to kinematics, the following equation can be obtained.

% \begin{equation}
% X(k+1)=X(k)+J(\Psi)\Delta\Psi(k+1)
% \end{equation}
% where $\Delta\Psi(k+1)$ represents the input of the robot system at time k+1. $J$ and $X$ represent the Jacobian matrix and end pose of the DABM, respectively.

%  \textcolor{red}{Reduce space before/after displayed equations}
% \setlength{\abovedisplayskip}{4pt}    % space above unbroken displays
% \setlength{\belowdisplayskip}{4pt}    % space below unbroken displays
% \setlength{\abovedisplayshortskip}{3pt}% space above after a short line
% \setlength{\belowdisplayshortskip}{3pt}% space below after a short line

% % If you use align/enumerate for multiple lines, also reduce the row separation:
% \setlength{\jot}{2pt}                 % line‐to‐line spacing in align environments

According to model prediction theory, under given target trajectory pose $X_{r}$ conditions, the optimal input $u$ of the system at each control moment can be obtained by solving the optimal solution of the objective function. The objective function and constraint conditions are as follows:
\begin{equation}
M\left (k \right ) =\sum_{i=k}^{N+k}e^{T}(k)Qe(k)+\sum_{i=k}^{N+k}\Delta u^{T}(k)P\Delta u(k),
\end{equation}
\begin{equation}
e(i) = X_{r}(k)-X(k), 
\end{equation}
\begin{equation}
X(k) = X(k-1)+J(u)\Delta u(k-1), 
\end{equation}
where $N$ is the predicted time domain, $J$ represent the Jacobian matrix, $\Delta u$ is the change in system input, $Q$ and $P$ are diagonal weighing matrices, the optimal solution of the objective function can be obtained to achieve the current system input $u$ and minimize pose error.

\subsection{Data-Driven Inverse-Kinematics Networks}
An alternative approach employs purely data-driven networks that learn a direct mapping from measured tip errors to actuator commands. Three architectures are examined in this work: feed-forward neural networks (FNN), long short-term memory (LSTM) networks, and gated recurrent units (GRU).

\subsubsection{Feed-forward neural network (FNN)}
A feedforward neural network (FNN) is a static, memoryless model that learns to directly map recent state histories to current control commands \cite{wang2024using, wang2021feedforward}. In this work, the FNN takes as input a fixed window of the last \(N\) tip error vectors for both continuum segments 
\begin{equation}
\mathbf E_{k-N:k-1} =
\left[
  \mathbf e(k-1)^{\!\top},\dots,\mathbf e(k-N)^{\!\top}
\right]^{\!\top}
\in \mathbb R^{6N},
\end{equation}
where each \(\mathbf{e}(i)\in\mathbb{R}^6\) concatenates the positions and orientation errors of dual-segments at time \(i\), and outputs the required actuator increments for each segment at the current time step 
\begin{equation}
    \Delta\boldsymbol\theta(k)=
\mathcal F_{\text{FNN}}\!\bigl(\mathbf E_{k-N:k-1}\bigr)
\in \mathbb R^{4},
\end{equation}
which indicates the vector of increases in motor angle for the four tendon actuators (two per segment) at time \(k\).
\subsubsection{LSTM network}
To capture long-range, nonlinear interactions, the standard LSTM cell \cite{wu2021hysteresis} with input, forget and output gates—stacked in 4 layers of 128 units, was implemented. In the proposed LSTM controller, an input sequence of the past robot states,
\begin{equation}
S_{t-T:t-1} = \{ X(t-T), u(t-T), \dots, X(t-1), u(t-1) \},
\end{equation}
is used to predict the motor rotation counts at the time $t$, where $X(t)$ denotes the six-dimensional pose (position and orientation) of the DABM tip and $u(t)$ the corresponding motor commands.

\subsubsection{GRU Controller}
\begin{figure}
  \centering
  \includegraphics[width=0.40\textwidth]{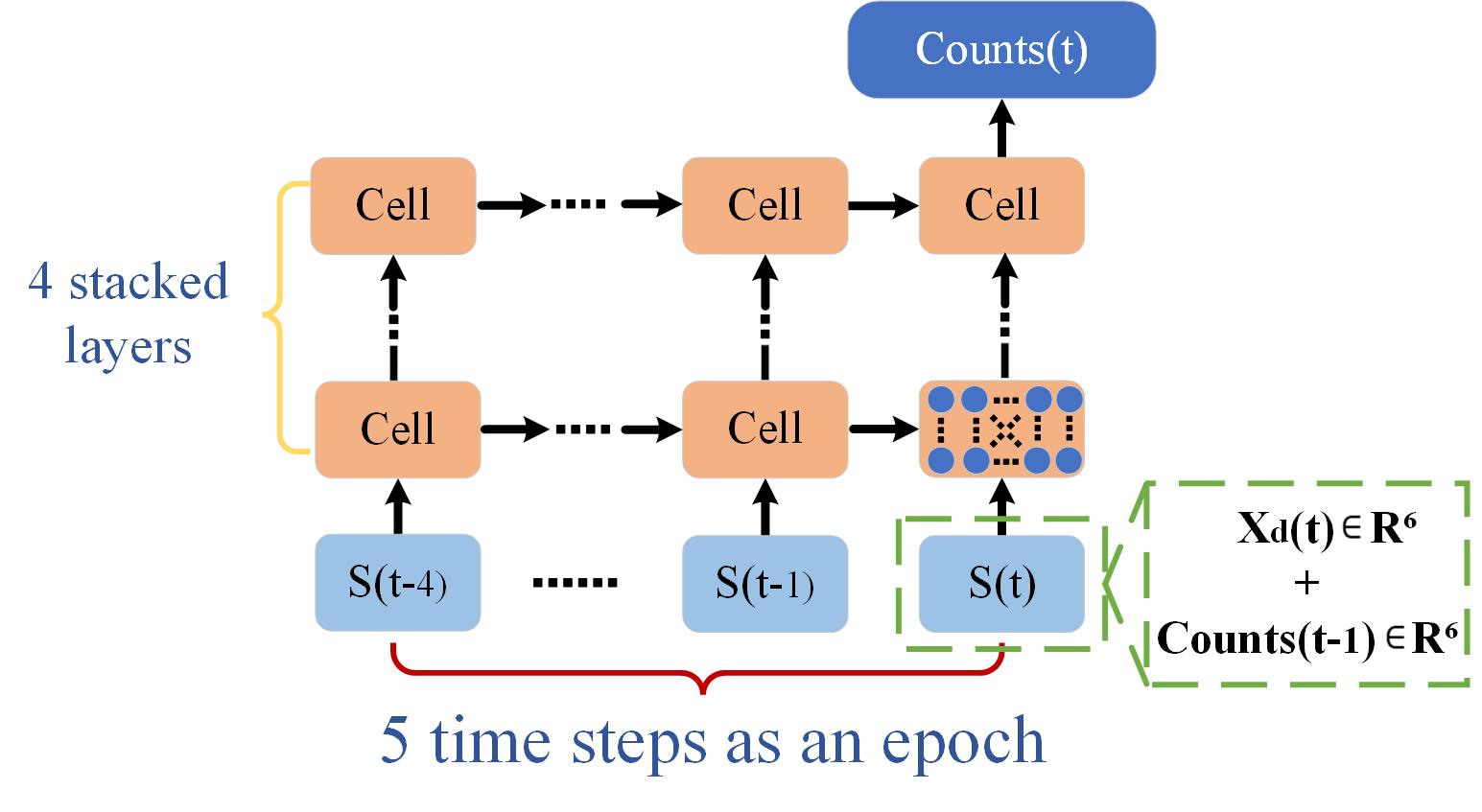} % need to be distinguished from the LSTM version 
  \caption{Internal GRU neural network structure for robot control: a 4 stacked layers GRU that is composed of multiple GRU cells. Five time steps are taken as the input epoch, and each time step state $S$ size is 12, which contains the target pose at the moment $t$ and the actual motor rotation counts from the previous moment $t-1$. In this work, each cell consists of 128 neurons, and the overall network is used to predict the motor rotation counts $(t)$.}
  \label{GRUCell}
\end{figure}

The gated recurrent unit (GRU) is a memory-based neural network that captures the temporal characteristics of robot dynamics and is adopted here for DESectBot pose control. Figure \ref{GRUCell} depicts the computations performed inside a single GRU cell \cite{schafke2024learning, yao2023rnn}:
\begin{equation}
\begin{aligned}
z(t) &= \sigma (W_z S(t) + U_z h(t-1) + b_z), \\
r(t) &= \sigma (W_r S(t) + U_r h(t-1) + b_r), \\
\tilde{h}(t) &= \tanh (W_h S(t) + U_h (r(t) \odot h(t - 1)) + b_h), \\
h(t) &= (1 - z(t)) \odot h(t - 1) + z(t) \odot \tilde{h}(t).
\end{aligned}
\end{equation}
where $S(t), h(t), z(t), r(t), \tilde{h}(t)$ represent the input vector, output vector, update gate vector, reset gate vector, and candidate activation vector for this unit. $W_*, U_*,$ and $b_*$ represent input weight, output weight, and bias parameters for each vector, which will be learned during training.

%% Method Comparison
\begin{table*}[ht]
  \centering
  \scriptsize % 可选：根据需要保留或删除
  \caption{Comparison of Control Methods for Two-Segment Continuum Robots}
  \label{tab:controller_comparison}
  \begin{tabularx}{1.0\textwidth}{@{\extracolsep{\fill}} l c c c X X}
    \toprule
    Controller & Method Type & Memory & Dual-Segment Mechanism Support & Advantages & Disadvantages \\
    \midrule
    Jacobian 
      & Model-based
      & None 
      & Basic: per-segment
      & Low latency; 
      & Sensitive to theoretical model error \\
    MPC 
      & Model-based
      & None 
      & Medium: includes multi-segment constraints 
      & Anticipates future states; 
      & Limited by model fidelity \\
    FNN 
      & Data-driven
      & None 
      & Low: no temporal memory; coupling only via data 
      & Fast inference and training 
      & Static mapping; lacks adaptivity \\
    LSTM 
      & Data-driven
      & Long-term 
      & High: learns inter-segment coupling 
      & Captures long-term dependencies;
      & Slower inference \\
    GRU 
      & Data-driven
      & Mid-term 
      & High: coupling learning with reduced complexity 
      & Lightweight; faster inference 
      & Slightly reduced memory vs. LSTM \\
    \bottomrule
  \end{tabularx}
  \vspace{-0.2in}
\end{table*}

%%  Hyperparameter tuning of the GRU
% define a new type table：table element水平＋垂直居中
\newcolumntype{C}[1]{>{\centering\arraybackslash}m{#1}}

\begin{table}[ht]
  \centering
  \caption{Hyperparameter tuning of GRU and LSTM structure}
  \label{table:gru_hyperparam_analysis}
  \renewcommand{\arraystretch}{0.80}
{\large
  \resizebox{\columnwidth}{!}{%
    \begin{tabular}{cccc}
      \toprule
      \textbf{Hidden layers number / Structure} 
        & \textbf{Hidden size} 
        & \textbf{Validation loss} 
        & \textbf{Training time per epoch (s)} \\
      \midrule
      3 / GRU &  64      & 0.1164                          & 3.6331 \\
      3 / LSTM &  64      &  0.1380                         & 5.1216 \\ 
      4 / GRU & 128      & \cellcolor{gray!20}\textbf{0.0501} & 5.0039 \\
      4 / LSTM & 128    &  0.1092                  &  7.0134  \\
      5 / GRU & 256      & 0.0507                          & 5.2529 \\
      5 / LSTM & 256    &   0.1165                 &  7.4827  \\
      \bottomrule
    \end{tabular}%
  }}
\end{table}

During training, the GRU receives the 12-dimensional input vector
\begin{equation}
\mathbf{S}_\ast(t)=
\bigl[
  \mathbf{X}_d(t)^{\!\top},\;
  \mathbf{u}(t-1)^{\!\top}
\bigr]^{\!\top}
\in \mathbb{R}^{12},
\end{equation}
which concatenates the six DoFs tip pose $\mathbf{X}_d(t)$ with the six motor-rotation counts $\mathbf{u}(t-1)$. The learning target is the current motor-rotation vector
\begin{equation}
\mathbf{h}_\ast(t)=\mathbf{u}(t)\in\mathbb{R}^{6}.
\end{equation}
At run time the pose term $\mathbf{X}_d(t)$ is replaced by the desired tip pose, following the scheme described in \cite{chen2023hybrid}.

\begin{figure}
  \centering
  \includegraphics[width=0.47\textwidth]{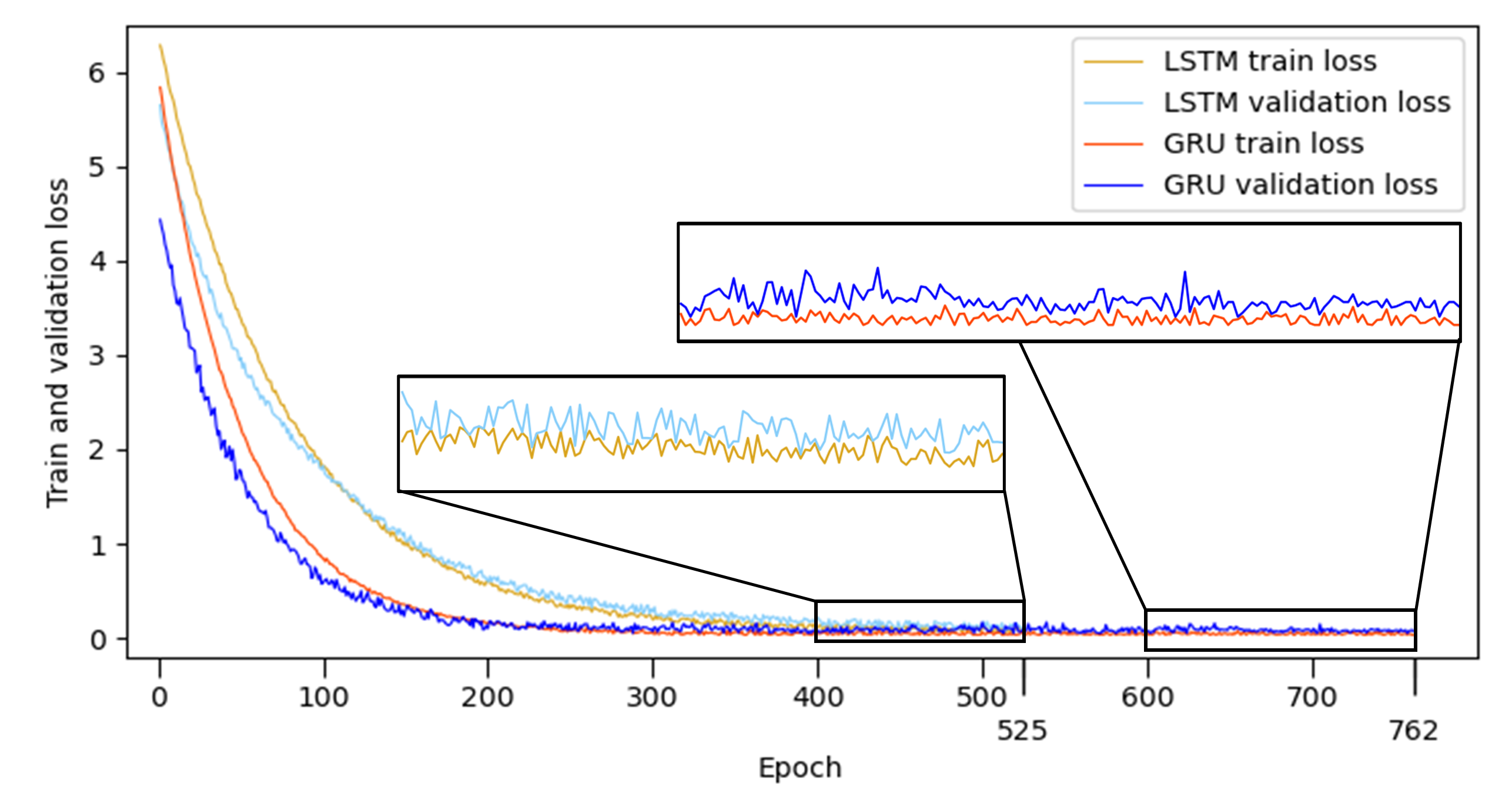} % need to be distinguished from the LSTM version 
  \caption{ Training and validation loss curves of GRU and LSTM models as a function of training epochs.}
  \label{LossPlot}
\end{figure}

% 定义一个新列类型 C，既水平也垂直居中，需要指定列宽 #1
\newcolumntype{C}[1]{>{\centering\arraybackslash}m{#1}}

% Optional：增加行高，让垂直居中更明显
\setlength{\extrarowheight}{3pt}

% \begin{table}[ht]
%   \centering
%   \caption{Hyperparameter tuning of GRU structure}
%   \label{tab:hyperparam-gru}
%   \begin{tabular}{%
%     C{4cm}  % 第一列宽 4cm
%     C{2cm}  
%     C{2cm}  
%     C{3cm}} 
%     \toprule
%     Number of hidden layers or batch size
%       & Hidden size
%       & Validation loss
%       & Training time per epoch (s) \\
%     \midrule
%     3   & 64  & 0.1164 & 3.6331 \\
%     4   & 128 & 0.0501 & 5.0039 \\
%     5   & 256 & 0.0507 & 5.2529 \\
%     \bottomrule
%   \end{tabular}
% \end{table}

% For GRU training, we employ robot poses $X_d$ (size:6) and actuation module six motors previous rotation counts(t-1) (size:6) as input $S_* \in R^{12}$, and current motor counts(t) (size:6) as output $h_* \in R^6$. To apply such a network as a controller, we replace the robot pose in the last step with the desired pose, following the strategy in \cite{chen2023hybrid}. The hyperparameters of the neural networks are shown in Table \ref{table:gru_parameter}.

\begin{table}[htbp]
\centering
\caption{GRU \& LSTM HYPERPARAMETER}
\label{table:gru_parameter}
\renewcommand{\arraystretch}{0.80}
\scriptsize
\begin{tabular}{ll|ll}
\toprule
\textbf{Parameter}      & \textbf{Value} & \textbf{Parameter}       & \textbf{Value} \\
\midrule
Layer Number            & 4            & Input Size               & 12           \\
Output Size             & 6            & Previous Time Step       & 5            \\
Hidden State Size       & 128          & Batch Size               & 64           \\                    
Optimizer               & Adam         & Learning Rate            & 0.001        \\
Loss Function  & L2 Loss & Epochs Number    & 1500             \\
\bottomrule
\end{tabular}

\end{table}

\vspace{-0.1in}
\subsection{Data Acquisition and Training}
\begin{figure}[hbt]
  \centering
  \includegraphics[width=0.49\textwidth]{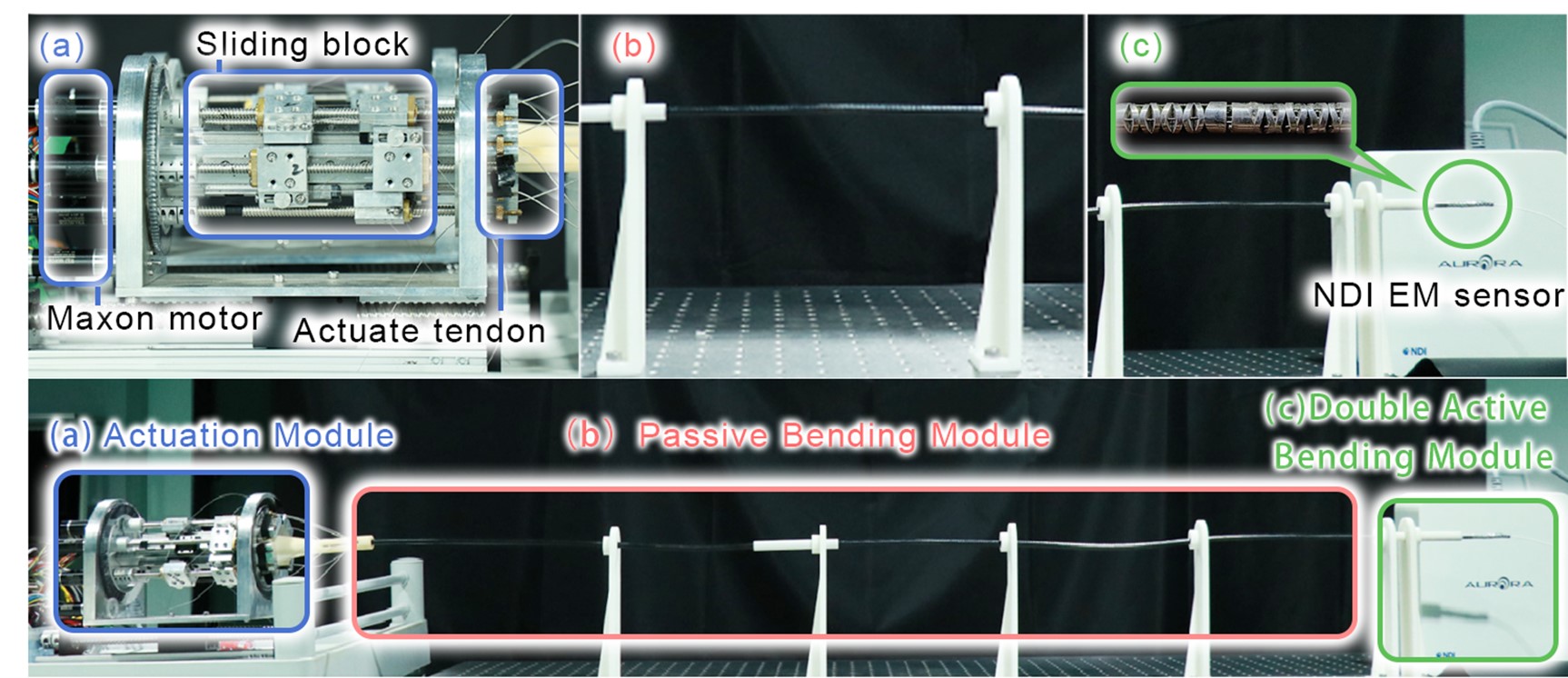}
  \caption{Experimental hardware setup for data acquisition, position, and orientation control. $Bottom $: (a) the DESectBot actuation module, (b) passive bending module with four support brackets to maintain its horizontal orientation, (c) double active bending module. $Top $: all parts are local close-up enlargements of the $Bottom$ section. (a) shows detailed structure of actuation module, (b) intercepts a section of the passive bending module and zooms in on the details, (c) shows the detailed structure of DABM and the white square object is the NDI$^\text{\textregistered}$ EM sensor equipment which its tracker tip is affixed to the central hole of the tool connection segment within the DABM.}
  \label{setup}
\end{figure}

The hardware experimental setup comprises three main components: the DESectBot platform, support brackets, and a 6 DoFs NDI$^\text{\textregistered}$ EM tracker. As depicted in Fig. \ref{setup}, four support brackets were installed to maintain the horizontal orientation of the DESectBot passive bending module. This setup accounts for the length of the module and natural sagging, ensuring smooth transmission of the actuation force from the actuation module to the DABM. Furthermore, the NDI$^\text{\textregistered}$ EM sensor tracker tip is affixed to the central hole of the tool connection segment within the DABM, with its predefined local coordinate system.

% Add another brief sensitivity analysis with Table II
The input and output mapping for the acquisition of GRU data are the DABM tip pose and its corresponding motor rotation counts. The general samples to train the GRU model are obtained using a Monte Carlo method, all training data trajectories are randomly generated and mapped with motor rotation counts on each time pin within a 4080 second duration, ensuring that the DESectBot can obtain sufficient training data for various situations to deal with multiple operable tasks and fall within a single safe working time at the same time. 
The NDI sensor measures position and orientation with respect to the local coordinate system at a rate of $5$ Hz, summing to a sample size of 20400 and its training, validation and test sets are 14280, 4080 and 2040 respectively. All data-driven models were trained on the identical dataset using same computing hardware condition (8 GB NVIDIA$^\text{\textregistered}$ RTX 4060Ti GPU).

To determine whether the observed performance advantage of the GRU arises from architectural suitability rather than hyperparameter bias, a compact ablation study was conducted by systematically varying the recurrent architecture (GRU vs. LSTM), the number of hidden units {64,128,256}, and the number of layers {3,4,5} under identical training conditions. All models shared the same input–output dimensions, number of time steps, optimizer, learning rate, and batch size. Early stopping was applied based on the validation MAE with a patience of 10 epochs.

As summarized in Table \ref{table:gru_hyperparam_analysis}, the GRU consistently achieved lower validation loss than the LSTM across all tested configurations. The best-performing model was a four-layer GRU with 128 hidden units, which yielded the lowest validation loss of 0.0501 while maintaining favorable computational efficiency. Shallower GRU models exhibited insufficient representational capacity, whereas deeper or wider configurations did not provide additional performance gains, indicating an optimal bias–variance trade-off at moderate depth. The LSTM followed a similar trend, with its best performance also achieved using four layers and 128 hidden units (validation loss 0.1092). However, LSTM models consistently exhibited higher validation loss and longer per-epoch training times, and increasing network depth or hidden size did not improve generalization. Fig. \ref{LossPlot} compares the learning curves of the best-performing GRU and LSTM models. While both architectures converge stably, the LSTM shows slower convergence, a larger generalization gap, and earlier termination due to early stopping (epoch 525), whereas the GRU continues improving until epoch 762. The complete set of evaluated hyperparameters is summarized in Table \ref{table:gru_parameter}. Overall, these results indicate that for the short-horizon temporal dynamics considered in this study, the GRU architecture provides a more effective balance between model capacity, generalization performance, and computational efficiency.
All tip-tracked data collected during the Monte Carlo sampling process are confined within the designed dome-shaped DABM workspace of $\phi$ $60 \times 45$ $mm$ (circular area × height), which defines the coverage of the training dataset. The evaluation trajectories (Fig. \ref{TaskA}) are defined within a smaller sub-region of $40\times 30$ $mm$ (outermost edge)  and two 15 mm-amplitude loop Lissajous curves which are entirely excluded from the training data.

% \begin{figure}
%   \centering
%   \includegraphics[width=0.3\textwidth]{Figures/workspace.png} 
%   \caption{\fcolorbox{gray}{yellow}{R2.1}\hl{Training data collection in the designed dome-shaped DABM workspace with a dimension of $\phi$ 60 $\times$ 45 mm (circle area $\times$ height)}}
%   \label{dataCollection}
% \end{figure}

%\input{Sections/Hysteresis_compensation}

\section{Experiments}

% \vspace{-0.5em} 

% In ESD procedures as mentioned above, the precision and dexterity of surgical instruments are highly important for surgery safety and success. To verify this, several motion control experiments were implemented in \textit{Task A and B}

To validate the proposed controller described in Section \ref{methods}, a benchtop experimental setup was established. Two kinds of trajectories and one set of orientation control experiments designed to closely mimic the clinical requirements of ESD procedures were implemented to assess the performance of the proposed GRU controller.
\vspace{-0.1in}

\subsection{Experimental setup and calibration}

% As shown in Fig.\ref{setup}, considering the length of DESectBot passive bending module and its hose that will naturally sag and flex, four support brackets were set up to maintain the passive bending module horizontal attitude. The NDI tracker tip is affixed to the central hole of the tool connection segment in active bending module, with its local coordinate system pre-defined. 

% To rigorously assess the performance of the proposed controllers under realistic surgical conditions, a comprehensive experimental setup was designed. This section details the hardware configuration and communication architecture used in the evaluation.

% The hardware platform for position and orientation control experiments closely follows the configuration used during data acquisition (see Section III). 

A comprehensive experimental setup—matching the hardware and communication architecture used during data acquisition (Section III)—was built to evaluate the controllers under realistic surgical conditions. In brief, the DESectBot system comprises:
\begin{itemize}
  \item A seven‐motor actuation module (Maxon ECXSP16L with GPX16HP gearboxes), split into five tendon‐drive motors, one rotation motor, and one translation motor. 
  \item A 1.2 m ebonite sheath serving as the passive bending module, supported by four custom brackets.
  \item A DABM with two 18 mm continuum segments constructed from spatial cross‐curved disks, each actuated by antagonistic NiTi tendons.
  \item A 6 DoFs electromagnetic tracker (NDI® Aurora V3.1) mounted on the DABM tip provides real‑time 5 Hz measurements of position and orientation within the robot’s workspace.
\end{itemize}

The Control software is hosted on a dedicated PC, and the model-based controller was developed in C++ and interfaces directly with the Maxon motor controllers through the manufacturer’s Controller Area Network-based API. This low‑level implementation achieves submillisecond command latency and deterministic cycle times.

In contrast, learning‐based data-driven controllers are developed in Python, leveraging PyTorch for neural network inference. To bridge between the Python environment and the C++ motor drivers and facilitate the testing of the algorithm model, a lightweight REST API using Flask was deployed:

\begin{itemize}
  \item \textbf{Flask server (Python):} Listens on TCP port for JSON-encoded control requests; decodes desired pose, runs the GRU forward pass, and returns motor increment commands.
  \item \textbf{C++ client:} Uses the standard Httplib library to asynchronously POST target poses to the Flask endpoint, parse the returned motor commands, and send them to the actuation module.
\end{itemize}

% Both controllers operate at 5 Hz and are synchronized by a shared Local Area Network (LAN) switch. TCP/IP communication is confined to a dedicated LAN to ensure isolation from other traffic and minimize jitter. An emergency stop handshake over a secondary UDP channel guarantees immediate motor shutdown within $2$ ms in the event of network failure or safety override.
Both controllers run at $5$ Hz, synchronized via a dedicated LAN switch. Control commands travel on an isolated TCP/IP network to minimize jitter, while a separate UDP link carries an emergency‑stop signal that shuts the motors down within $2$ ms if communication fails or safety is triggered. 
%%% inference time 
 The mean actual inference time (in milliseconds) for each controller (Jacobian 11.37 ms, MPC 17.45 ms, FNN 5.02 ms, LSTM 12.63ms, and GRU 8.91ms) on the deployment hardware, providing a practical comparison of computational efficiency.

% Considering that the DESectBot’s DABM pose drifts away from the initial zero reference after each experiment, the fixed position of the NDI planar field generator was established as the global reference frame to ensure identical starting configurations across repetitions. 

To standardize every trial, the fixed NDI planar field generator defines the global frame. After each run, an auto‑reset script zeroes motor counts, thereby returning the end effector close to its zero pose. The Elmo Application Studio II (64-bit) interface was used to increment each motor encoder in $\pm 500$‑count steps. Real‑time pose feedback from NDI ToolBox (v5.000.019) guided successive adjustments until the end‑effector error was reduced to within $\pm 0.5$ mm in position and $\pm 0.3^{\circ}$ in Euler orientation—values matching the sensor’s root mean square error (RMSE) accuracy ($0.48$ mm and $0.3^{\circ}$, respectively). All timestamps and poses are logged, a custom GUI streams live configuration and errors, and a model‑based optimizer is being developed to replace the current manual fine‑tuning.

\vspace{-0.1in}
\subsection{Position and orientation control}
\label{Task A}

% In this section, the five control strategies detailed in Sec. \ref{methods} are assessed with respect to the two kinematic objectives most pertinent to endoscopic surgery—Cartesian position regulation and end‑effector orientation control. The experimental results are presented in the subsections that follow.
% To emulate representative intra‑luminal manoeuvres, two canonical path‑tracking tasks were implemented. Marking and circumferential incision require the instrument tip to trace smooth, arc‑like curves, whereas submucosal dissection and tissue lifting demand rectilinear advancement along straight lines \cite{gao2024transendoscopic}. Accordingly, an arc‑based Lissajous pattern and a line‑based nested‑rectangle trajectory were selected as benchmark paths. Each controller was executed five times on both trajectories to quantify repeatability and robustness.

This section benchmarks the five controllers from Section \ref{methods} against the two kinematic goals most relevant to endoscopic work—precise Cartesian positioning and accurate end‑effector orientation.

To mimic typical intra‑luminal motions, two path‑tracking tasks were chosen. A Lissajous curve represents the smooth arcs used for marking and circumferential cutting, while a nested‑rectangle path captures the straight‑line advances required for submucosal dissection and tissue lifting \cite{gao2024transendoscopic}. Each controller performed five runs on both trajectories, allowing a fair comparison of accuracy, repeatability, and robustness.
% In this section, the five controllers described in Sec. \ref{methods} through two key aspects commonly encountered in clinical applications: position control and orientation control. The experimental results are presented as follows.

% %%% Emphasize on two EIGHT combination, more difficult
% Surgical instrument movement for marking and circumferential incision always need to follow the precise arc-based trajectories, whereas submucosal dissection and tissue lifting require straight, line‐based tool advancement \cite{gao2024transendoscopic}.
% Therefore, two types of trajectory tracking tasks were selected as : Nested rectangle (line-based) and Lissajous (arc-based). \hl{Moreover, to assess performance stability, each of the five controllers was tested five times on every trajectory.}

% In this section, we validated three kinds of controllers within two tasks that can only be achieved by segmented robots, the experimental process and results are demonstrated and analyzed as follows. 

\subsubsection{\textbf{Nested Rectangle}}
\begin{figure*}[htb]
  \centering
  \includegraphics[width=1.0\linewidth]{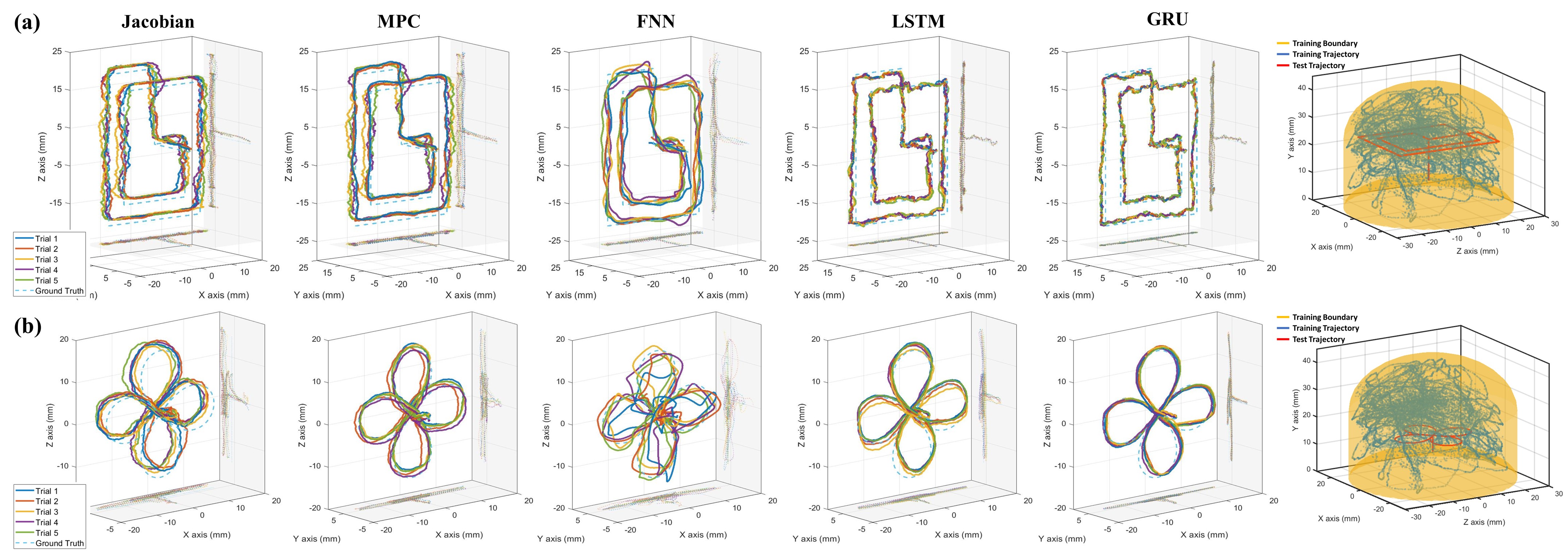}
  \caption{Trajectory experiment: (a) Three-dimensional visualization of the five trials for each controller motion trajectory with one ground truth nested rectangle trajectory, (b) Three-dimensional visualization of the five trials for each controller motion trajectory with one ground truth lissajous trajectory. The yellow designed dome-shaped volume indicates the training data collection workspace which has a dimension of $\phi$ $60 \times 45$ $mm$ (circular area × height), within which all training trajectories were sampled before. The red curve represents the experimental test trajectory, excluded from the training set and used solely for performance evaluation.}
  \label{TaskA}
  \vspace{-0.1in}
\end{figure*}

%%%%%%%%% Table for Task A %%%%%%%%%
\begin{table*}[!t]
  \centering
  \caption{DATA ANALYSIS OF TASK A}
  \label{tab:data-analysis-task-a-vertical}
  \scriptsize
  \setlength{\tabcolsep}{1pt}
  \renewcommand{\arraystretch}{0.7}
  \begin{tabular*}{0.70\textwidth}{@{\extracolsep{\fill}} l *{6}{c} cc}
    \toprule
    \multicolumn{9}{c}{\textbf{Nested Rectangle}} \\
    \midrule
    \textbf{Controller} 
      & \multicolumn{6}{c}{\textbf{MAE \textbar\ STD}} 
      & \multicolumn{2}{c}{\textbf{RMSE}} \\
    & Pos X (mm) & Pos Y (mm) & Pos Z (mm)
      & Ori X (°)   & Ori Y (°)   & Ori Z (°)
      & Pos(mm)   & Ori(°)\\
    \midrule
    Jacobian & 0.93 \textbar\  0.73& 1.47 \textbar\  0.62& 1.23 \textbar\ 0.43 & 3.14 \textbar\  1.59& 5.87 \textbar\  1.68& 4.34 \textbar\ 2.19 & 2.38 & 9.70 \\
    MPC      & 0.68 \textbar\ 0.14& 0.60 \textbar\ 0.17& 1.31 \textbar\ 0.26& \textbf{\hlnum{1.43}} \textbar\ 1.18 & 3.12 \textbar\ \textbf{\hlnum{0.82}}& 2.50 \textbar\ 1.96& 1.89 & 4.98(98) \\
    FNN      & 0.76 \textbar\ 0.69& 0.60 \textbar\ 0.63& 0.73 \textbar\ 0.64& 2.85 \textbar\ 2.05& 4.42 \textbar\ 1.94& 2.77 \textbar\ 2.97& 1.50 & 6.80 \\
    
    LSTM     & 0.60 \textbar\ 0.34& 0.47 \textbar\ 0.27& \textbf{\hlnum{0.61}} \textbar\ 0.33& 1.80 \textbar\ 1.26& 3.25 \textbar\ 1.36& 2.28 \textbar\ 1.98& 1.13 & 4.98(76)  \\
    
    GRU      & \cellcolor{gray!20}\textbf{0.57} \textbar\ \textbf{0.10}& \cellcolor{gray!20}\textbf{0.45} \textbar\ \textbf{0.11}& 0.62 \textbar\ \textbf{\hlnum{0.15}}& 1.59 \textbar\ \textbf{\hlnum{1.15}}& \textbf{\hlnum{2.99}} \textbar\ 1.05& \cellcolor{gray!20}\textbf{2.12} \textbar\ \textbf{1.17}& \cellcolor{gray!20}\textbf{1.11} & \cellcolor{gray!20}\textbf{4.62} \\
    \midrule
    \multicolumn{9}{c}{\textbf{Lissajous}} \\
    \midrule
    \textbf{Controller} 
      & \multicolumn{6}{c}{\textbf{MAE \textbar\ STD }} 
      & \multicolumn{2}{c}{\textbf{RMSE}} \\
    & Pos X (mm) & Pos Y (mm) & Pos Z (mm)
      & Ori X (°)   & Ori Y (°)   & Ori Z (°)
      & Pos    & Ori\\
    \midrule
    Jacobian & 1.00 \textbar\ 0.91& 1.22 \textbar\ 0.53& 0.96 \textbar\ 0.25& 4.74 \textbar\ 1.94& 6.34 \textbar\ 1.50& 3.81 \textbar\ 2.02& 1.61 & 6.86 \\
    MPC      & 0.87 \textbar\ 0.53& 0.64 \textbar\ 0.63& 0.84 \textbar\ 0.47& 0.52 \textbar\ 0.43& 3.21 \textbar\ 0.96& 3.11 \textbar\ 1.65& 1.19 & 3.80 \\
    FNN      & 1.02 \textbar\ 1.09& 0.78 \textbar\ 0.83& 1.05 \textbar\ 1.11& 2.39 \textbar\ 2.15& 4.00 \textbar\ 1.74& 2.68 \textbar\ 2.21& 1.58 & 4.52 \\
    LSTM     & 0.60 \textbar\ \textbf{\hlnum{0.20}}& 0.49 \textbar\ 0.38& 1.00 \textbar\ 0.55& 0.49 \textbar\ 0.36& 2.58 \textbar\ 0.38& 2.51(03) \textbar\ 0.48& 1.07 & 2.65  \\
    GRU      & \textbf{\hlnum{0.58}} \textbar\ 0.34& \cellcolor{gray!20}\textbf{0.38} \textbar\ \textbf{0.07}& \cellcolor{gray!20}\textbf{0.62} \textbar\ \textbf{0.19}& \cellcolor{gray!20}\textbf{0.42} \textbar\ \textbf{0.17}& \cellcolor{gray!20}\textbf{2.56} \textbar\ \textbf{0.28}& \cellcolor{gray!20}\textbf{2.50(61)} \textbar\ \textbf{0.24}& \cellcolor{gray!20}\textbf{0.81} & \cellcolor{gray!20}\textbf{2.59} \\
    \bottomrule
  \end{tabular*}
  \vspace{-0.2in}
\end{table*}
The nested‑rectangle path consists of concentric rectangles connected by straight segments. As shown in Fig. \ref{TaskA}(a), the tip first translates $15$ mm along the +Y axis to reach the target plane. It then completes a rectangular loop; after each lap, the rectangle is enlarged by $5$ mm on every side until the outermost loop measures 40 mm×30 mm. During tracking, the NDI sensor samples the tip pose at $5$ Hz, while the DABM keeps its orientation fixed along +Y.

%%%Modofied%%%%

% The first type of trajectory involves multiple nested rectangular paths based on straight lines. The trajectory begins with a distance $15$ mm straight line to the trajectory surface along the Y axis (Fig. \ref{Rectangle} (a)), in alignment with the natural elongation direction of the DABM. Upon reaching the surface, the tip follows a rectangular path, with each subsequent nested rectangle increasing by $5$ mm at the edges, culminating in a final rectangle with dimensions of $40$ mm in length and $30$ mm in width. During trajectory tracking, the DESectBot updates its position at a frequency of $5$ Hz, as monitored by the NDI sensor. Furthermore, the orientation of the DABM is consistently maintained along the positive Y-axis.

%%%%%%%%% Discussion for Rectangles

As shown in the Fig. \ref{TaskA} (a), the GRU‐based controller yields the tightest clustering around the ground truth path, indicating superior trial‐to‐trial consistency. Quantitatively, for the nested‐rectangle task, GRU attains MAE of 0.57 mm (X), 0.45 mm (Y) and 0.62 mm (Z) (Table \ref{tab:data-analysis-task-a-vertical}), representing reductions of roughly 5 \% in X and 4 \% in Y compared to LSTM (0.60 / 0.47 mm) and over 15 \% compared to MPC (0.68 / 0.60 mm). While the GRU attains a marginally higher MAE in the $Z$-direction ($0.62$ mm versus $0.61$ mm for the LSTM),  it comprises only 350208 trainable parameters—approximately 75\% of the LSTM’s 466944—thereby reducing computational overhead and yielding faster inference. Furthermore, the standard deviation (STD) of all position errors indicates the GRU enhanced robustness and consistency and can achieve more stable and repeatable tracking performance.\\
In orientation tracking the GRU posts MAEs of $1.59^{\circ}$ (X), $2.99^{\circ}$ (Y) and $2.12^{\circ}$ (Z), representing approximately 12 \%, 8\% and 7\% gains over LSTM ($1.80^{\circ}$/$3.25^{\circ}$/$2.28^{\circ}$). MPC does edge the GRU on two isolated metrics—orientation‑X MAE ($1.43^{\circ}$) and orientation‑Y STD ($0.82^{\circ}$)—yet the GRU prevails elsewhere, yielding the lowest positional errors on every axis, the smallest orientation‑Z spread, and the best overall RMSE ($RMSE_{pos} = 1.11$ mm, $RMSE_{ori} = 4.62^{\circ}$ mm). For comparison of RMSE, LSTM records 1.13 mm/$4.99^{\circ}$, MPC 1.89 mm/$4.99^{\circ}$, and the Jacobian 2.38 mm/$9.70^{\circ}$. Hence, across the nested rectangle task, the GRU offers the highest spatial precision and repeatability among all controllers evaluated.

% In orientation, GRU achieves MAEs of 1.59° (X), 2.99° (Y) and 2.12° (Z), improving by approximately 12\%, 8\% and 7\%, respectively, relative to LSTM (1.80°/ 3.25°/ 2.28°). \hl{Although MPC shows partial advantages in MAE and STD performance in a single orientation respectively}, the root mean square errors (RMSE) further underscore GRU’s advantage: $RMSE_{pos} = 1.11$ mm, $RMSE_{ori} = 4.62^{\circ}$ mm versus 1.13 mm/ 4.9876° for LSTM, 1.89 mm/ 4.9898 ° for MPC, and 2.38 mm/ 9.70° for the Jacobian method. These results confirm that the GRU controller delivers the highest spatial precision and repeatability among all tested methods. 

% \[
% \text{RMSE}_{\text{pos}} = 1.11~\text{mm},\quad \text{RMSE}_{\text{ori}} = 4.62^\circ,
% \]  
\vspace{-0.1in}
\subsubsection{\textbf{Lissajous}}

Because curved motions dominate endoscopic work and are harder to control than straight lines, all five controllers on a Lissajous path were tested \cite{gao2024transendoscopic}. The reference consists of two $15$ mm‑amplitude loops—one in Z, one in X—centered at the same point. Starting $6$ mm above the X–Z plane, the DABM tip completes the vertical loop first and then transitions smoothly into the horizontal loop. Throughout tracking, DESectBot updates its pose at $5$ Hz while keeping the end‑effector oriented along +Y.
As shown in Fig. \ref{TaskA} (b) and quantified in Table \ref{tab:data-analysis-task-a-vertical}, the GRU controller outperforms all other methods on the Lissajous path, posting the lowest values for both MAE and STD. It attains the lowest positional RMSE of $0.81\,$mm and orientational RMSE of $2.59^\circ$, outperforming LSTM (1.07 mm/ 2.65°), MPC (1.19 mm/ 3.80°) and the Jacobian baseline (1.61 mm/ 6.86°). These results demonstrate that the GRU adaptation effectively compensates for nonlinear coupling in the dual‐segment continuum, yielding superior tracking fidelity under clinically realistic curved trajectories.  

\begin{table*}[!t]
  \centering
  \caption{DATA ANALYSIS OF TASK B}
  \label{tab:data-analysis-task-b}
  \scriptsize
  \setlength{\tabcolsep}{1pt}
  \renewcommand{\arraystretch}{0.78}
  \begin{tabular*}{0.73\textwidth}{@{\extracolsep{\fill}}
      >{\raggedleft\arraybackslash}p{2.5em}@{\quad\quad}l
      *{6}{c} cc}
    \toprule
    \multirow{2}{*}{Tip Points} & \multirow{2}{*}{Controller}
      & \multicolumn{6}{c}{\textbf{MAE \textbar\ STD}} 
      & \multicolumn{2}{c}{\textbf{RMSE}} \\
    &  & Pos X (mm) & Pos Y (mm) & Pos Z (mm)
         & Ori X (°)   & Ori Y (°)   & Ori Z (°)
      & Pos(mm)    & Ori(°) \\
    \midrule
    \multirow[c]{5}{*}{P1}
      & Jacobian & 0.79 \textbar\ 0.16& 0.68 \textbar\ 0.35& 1.40 \textbar\ 0.29& 5.98 \textbar\ 1.52& 5.59 \textbar\ 1.43& 6.57 \textbar\ 1.49& 0.54 &  3.12 \\
      & MPC      & 0.45 \textbar\ 0.08& 0.41 \textbar\ 0.15& 0.55 \textbar\ 0.13& 2.80 \textbar\ 0.53& 3.05 \textbar\ 0.69& 3.49 \textbar\ 0.75& 0.28 & 1.64  \\
      & FNN      & 0.46 \textbar\ 0.12 & 0.58 \textbar\ 0.17& 0.39 \textbar\ 0.15& 3.35 \textbar\ 1.03& 2.80 \textbar\ 1.27& 3.52 \textbar\ 1.18& 0.28 & 1.86  \\
      & LSTM     & 0.24 \textbar\ 0.06& 0.33 \textbar\ 0.08& 0.21 \textbar\ \textbf{\hlnum{0.06}}& 1.83 \textbar\  0.46& 1.40 \textbar\ 0.39& 1.95 \textbar\ 0.48& 0.15 & 1.01 \\
      & GRU      & \cellcolor{gray!20}\textbf{0.23} \textbar\ \textbf{0.04}
                 & \cellcolor{gray!20}\textbf{0.32} \textbar\ \textbf{0.07}
                 & \textbf{\hlnum{0.18}} \textbar\ 0.08
                 & \cellcolor{gray!20}\textbf{1.66} \textbar\ \textbf{0.34}
                 & \cellcolor{gray!20}\textbf{1.26} \textbar\ \textbf{0.30}
                 & \cellcolor{gray!20}\textbf{1.78} \textbar\ \textbf{0.37}
                 & \cellcolor{gray!20}\textbf{0.14} 
                 & \cellcolor{gray!20}\textbf{0.92}  \\
    \midrule
    \multirow[c]{5}{*}{P2}
      & Jacobian & 1.20 \textbar\ 0.28& 0.82 \textbar\ 0.34& 1.24 \textbar\ 0.40& 6.31 \textbar\ 1.61& 6.53 \textbar\ 1.20& 6.75 \textbar\ 1.65& 0.65 & 3.32  \\
      & MPC      & 0.41 \textbar\ 0.09& 0.30 \textbar\ \textbf{\hlnum{0.05}}& 0.46 \textbar\ 0.09& 2.16 \textbar\ 0.45& 2.12 \textbar\ 0.63& 2.70 \textbar\ 0.55& 0.17 &  1.20 \\
      & FNN      & 0.65 \textbar\ 0.39& 0.60 \textbar\ 0.45& 0.63 \textbar\ 0.47 & 2.06 \textbar\ 0.69& 3.26 \textbar\ 0.75& 2.08 \textbar\ 0.71& 0.73 &  1.21 \\
      & LSTM     & 0.31 \textbar\ 0.05& 0.21 \textbar\ 0.09& 0.25 \textbar\ 0.20& 1.16 \textbar\ 0.31& 1.24 \textbar\ 0.52& 1.60 \textbar\ 0.30& 0.15 & 0.83 \\
      & GRU      & \cellcolor{gray!20}\textbf{0.27} \textbar\ \textbf{0.03}
                 & \textbf{\hlnum{0.18}} \textbar\ \textbf{0.08}
                 & \cellcolor{gray!20}\textbf{0.22} \textbar\ \textbf{0.02}
                 & \cellcolor{gray!20}\textbf{0.99} \textbar\ \textbf{0.28}
                 & \cellcolor{gray!20}\textbf{1.07} \textbar\ \textbf{0.30}
                 & \cellcolor{gray!20}\textbf{1.38} \textbar\ \textbf{0.23}
                 & \cellcolor{gray!20}\textbf{0.13} 
                 & \cellcolor{gray!20}\textbf{0.71} \\
    \midrule
    \multirow[c]{5}{*}{P3}
      & Jacobian & 0.95 \textbar\ 0.23& 0.85 \textbar\ 0.28& 0.65 \textbar\ 0.33& 6.64 \textbar\ 1.29& 6.12 \textbar\ 1.47& 6.79 \textbar\ 1.36& 0.44 & 3.43  \\
      & MPC      & 0.52 \textbar\ 0.17& 0.30 \textbar\ 0.22& 0.51 \textbar\ 0.25& 2.43 \textbar\ 0.61& 2.77 \textbar\ 0.68 & 4.30 \textbar\  0.74& 0.26 & 1.77  \\
      & FNN      & 0.23 \textbar\ 0.33 & 1.04 \textbar\ 0.25& 1.43 \textbar\ 0.29& 1.80 \textbar\ 0.79& 3.37 \textbar\ 1.02& 5.24 \textbar\ 1.03& 0.52 & 2.12  \\
      & LSTM     & 0.20 \textbar\ 0.11& 0.38 \textbar\ 0.10& 0.45 \textbar\ 0.07& 0.96 \textbar\ 0.40& 1.75 \textbar\ 0.65& 1.95 \textbar\ 0.52& 0.18 & 1.17  \\
      & GRU      & \cellcolor{gray!20}\textbf{0.17} \textbar\ \textbf{0.07}
                 & \cellcolor{gray!20}\textbf{0.26} \textbar\ \textbf{0.06}
                 & \cellcolor{gray!20}\textbf{0.26} \textbar\ \textbf{0.02}
                 & \cellcolor{gray!20}\textbf{0.70} \textbar\ \textbf{0.13}
                 & \cellcolor{gray!20}\textbf{0.78} \textbar\ \textbf{0.32}
                 & \cellcolor{gray!20}\textbf{1.20} \textbar\ \textbf{0.29}
                 & \cellcolor{gray!20}\textbf{0.12} 
                 & \cellcolor{gray!20}\textbf{0.48} \\
    \midrule
    \multirow[c]{5}{*}{P4}
      & Jacobian & 0.75 \textbar\ 0.19& 0.39 \textbar\ 0.23& 0.22 \textbar\ 0.25& 5.15 \textbar\ 1.08& 4.70 \textbar\ 1.15& 5.01 \textbar\ 1.19& 0.33 & 2.68  \\
      & MPC      & 0.50 \textbar\ 0.28& 0.35 \textbar\ 0.11& 0.47 \textbar\ 0.12& 2.26 \textbar\ 0.37& 2.25 \textbar\ 0.53& 2.50 \textbar\ 0.49& 0.25 & 1.27  \\
      & FNN      & 0.54 \textbar\ 0.15& 0.28 \textbar\ 0.12& 0.28 \textbar\ 0.15& 1.42 \textbar\ 0.49& 1.69 \textbar\ 0.69& 2.01 \textbar\ 0.65& 0.21 & 1.01   \\
      & LSTM     & 0.49 \textbar\ \textbf{\hlnum{0.07}}& 0.18 \textbar\ 0.10& 0.25 \textbar\ \textbf{\hlnum{0.05}}& 1.17 \textbar\ 0.38& 1.39 \textbar\ 0.57& 1.66 \textbar\ \textbf{\hlnum{0.28}}& 0.18  & 0.83 \\
      & GRU      & \textbf{\hlnum{0.45}} \textbar\ 0.09
                 & \cellcolor{gray!20}\textbf{0.17} \textbar\ \textbf{0.07}
                 & \textbf{\hlnum{0.23}} \textbar\ 0.06
                 & \cellcolor{gray!20}\textbf{1.10} \textbar\ \textbf{0.26}
                 & \cellcolor{gray!20}\textbf{1.29} \textbar\ \textbf{0.44}
                 & \textbf{\hlnum{1.54}} \textbar\ 0.35
                 & \cellcolor{gray!20}\textbf{0.17} 
                 & \cellcolor{gray!20}\textbf{0.77}  \\
    \bottomrule
  \end{tabular*}
  \vspace{-0.1in}
\end{table*}

\vspace{-0.1in}
\subsection{Orientation control with fixed tip position}
\label{Task B}

\begin{figure}
  \centering
  \includegraphics[width=0.42\textwidth]{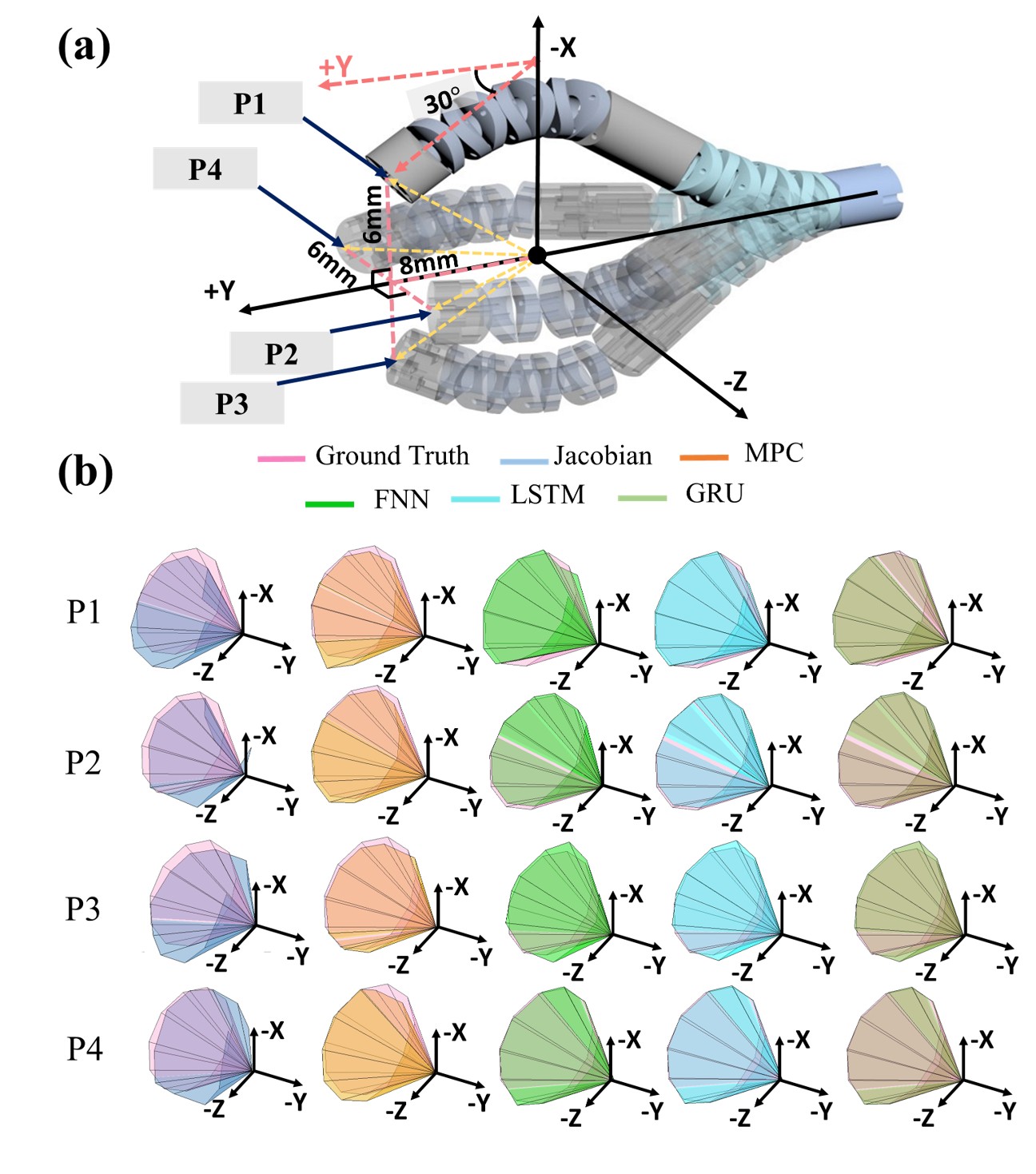}
  %{Figures/coneWithOrientationWithOrganPhantom.jpg}
  \caption{(a) The DABM configuration for orientation control with four fixed tip positions P1-P4, (b) Reconstruction of orientation control experiment: a swept area that extends into a cone with a vertex angle of $30^{\circ}$.}
  \label{OrientReconstrct}
  \vspace{-0.15in}
\end{figure}
% In actual ESD procedures, surgeons are often required to perform complex maneuvers within their confined workspace. Dexterity and real-time orientation control of the surgical instruments are key to the success of the procedure once the designated lesion has been reached.

% Due to the constrained workspace and dynamic environment, real-time orientation control of these instruments becomes crucial.

% In actual ESD procedures, maintaining the tip position of surgical instruments at the designated lesion is often required, the orientation control  is essential for successfully performing complex maneuvers, enabling surgeons to adapt effectively within the confined work space and dynamic environment\cite{rice2017esophagus}. Thus, in this section, experiments were designed to adjust the orientation of the DABM while maintaining the fixed position of the DESectBot's tip. 
In ESD the instrument tip must stay anchored on the lesion while its orientation is repeatedly adjusted to execute complex maneuvers within a cramped, shifting workspace \cite{rice2017esophagus}. To evaluate this capability, experiments that rotate the DABM about a fixed DESectBot tip position were conducted. As depicted in Fig. \ref{OrientReconstrct} (a), four fixed tip positions of the DABM are defined and labeled as:         
$P1$ $(-6,8,0)$; $P2$ $(0,8,-6)$; $P3$ $(6,8,0)$; $P4$ $(0,8,6)$.\par
At each position, the orientation of the DABM tip is set at an angle of $30^{\circ}$ relative to the $Y$-axis. Subsequently, the position of the DABM tip remains fixed, while its orientation rotates incrementally by $30^{\circ}$ around the $Y$-axis. This rotation results in a swept area that extends into a cone with a vertex angle of $30^\circ$.

Figure \ref{OrientReconstrct} (b) presents a comparison between the ground truth and the average orientation reconstruction (all five controllers was repeated five times orientation control on each point). To clearly demonstrate the difference between the ground truth and the performance of the five controllers in each frame, every $30^\circ$ step is taken as one sample for reconstruction. Table \ref{tab:data-analysis-task-b} illustrates the MAE, STD and RMSE for positions (mm) and orientation ($^\circ$) , after combining it with Figure \ref{OrientReconstrct} (b), it is evident that in all four position scenarios, the overlap between the ground truth (pink) and the LSTM (cyan) and GRU controller (green) surfaces are the most significant compared to the other controllers. The  quantitative analysis illustrates that the GRU controller markedly outperforms both model‐based and alternative data‐driven approaches in terms of both positional and orientational accuracy for all tests in MAE. In addition, for the vast majority of pose cases, the GRU achieved the minimum STD, demonstrating its excellent robustness in orientation control. Averaged across the four test points (P1–P4), GRU achieves the mean RMSE of 0.14 mm in position and 0.72° in orientation, compared with 0.17 mm/ 0.96° for LSTM, 0.435 mm/ 1.24° for FNN, 0.24 mm/ 1.46° for MPC, and 0.49 mm/ 3.14° for the Jacobian method.
Notably, the GRU controller reduces positional RMSE by approximately 18 \% relative to LSTM and by 36 \% relative to MPC, while orientation RMSE is lowered by 25 \% and 51 \%, respectively. 
% Mean absolute errors corroborate these findings: GRU maintains MAE $\leq 0.28$ mm along each Cartesian axis and $\leq 1.29$° in any Euler rotation, compared with MAE values up to 0.93 mm/ 6.16° for the Jacobian baseline.

% These results indicate that the GRU can make inference to nonlinear, time‐dependent coupling between the two continuum segments yields the highest spatial precision and repeatability. Consequently, the GRU integrated DESectBot is exceptionally well‐suited for the sub‐millimetric positioning and fine orientational adjustments required in clinical ESD applications and also confirming its superior kinematic registration and control precision in two-segment continuum robot.

%% 7.24 Modified paragragh
The results demonstrate that the GRU effectively models the nonlinear, time‑varying coupling between the two continuum segments, achieving the highest spatial accuracy and repeatability. Consequently, a GRU‑driven DESectBot provides sub‑millimetric positioning and precise orientation control, making it well suited for clinical ESD tasks and confirming its superior kinematic registration and control in a two‑segment continuum robot.

\section{Peg Transfer Benchmark and \textit{Ex Vivo} Evaluation}
%%% PEG TRANSFER %%%
As demonstrated in Section IV, the proposed GRU controller delivers superior kinematic registration and control accuracy in both position and orientation. Consequently, the DESectBot equipped with the GRU algorithm will be employed for the subsequent surgical task evaluations.\par
The peg‑transfer task is a well‑established, standardized exercise within surgical skills curricula \cite{derossis1998development} and is frequently adopted in endoscopic submucosal dissection (ESD) studies for benchmarking and training purposes \cite{habaz2019adaptation}. In the present work, a unilateral peg‑transfer protocol—executed with a single robotic arm—was implemented. Following the definitions in \cite{seung2015single}-\cite{hwang2022automating}: 
\begin{itemize}
\item \textbf{Pick:} the DESectBot grasps a block and lifts it clear of its initial peg.
\item \textbf{Transferring:} while maintaining the grasp, the block is conveyed to the designated target peg.
\item \textbf{Place:} the block is accurately deposited onto the target peg.
\end{itemize}

\begin{figure}[hbt]
  \centering
\includegraphics[width=0.5\textwidth]{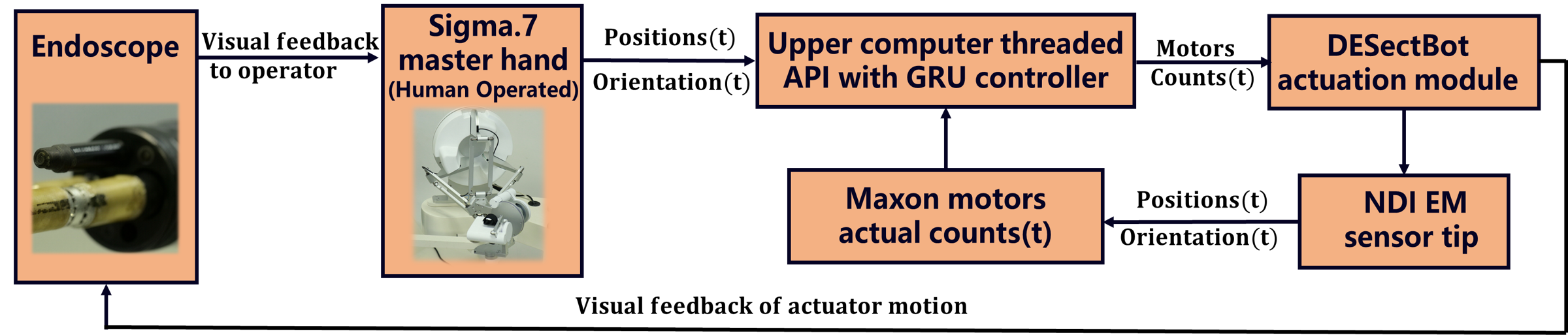}
  \caption{The brief teleoperation scheme of DESecBot alongside an endoscope with Sigma.7 master hand , the Motors Counts(t) represents this time step predicted motor counts for actuation module.}
  \label{Controlflowchart}
\end{figure}

 Compared with the aforementioned experiments, this test was utilized as a real-time input rather than predetermined trajectories. The pose of the DABM tip in each frame was controlled by a 6 DoFs master haptic interface (Sigma.7, Force Dimension AG, Switzerland), allowing for real-time adjustments. The DESectBot teleoperation scheme is shown in Fig. \ref{Controlflowchart}. 

  % Peg transfer control scheme

% Peg Tranjectories
\begin{figure}[hbt]
  \centering
\includegraphics[width=0.35\textwidth]{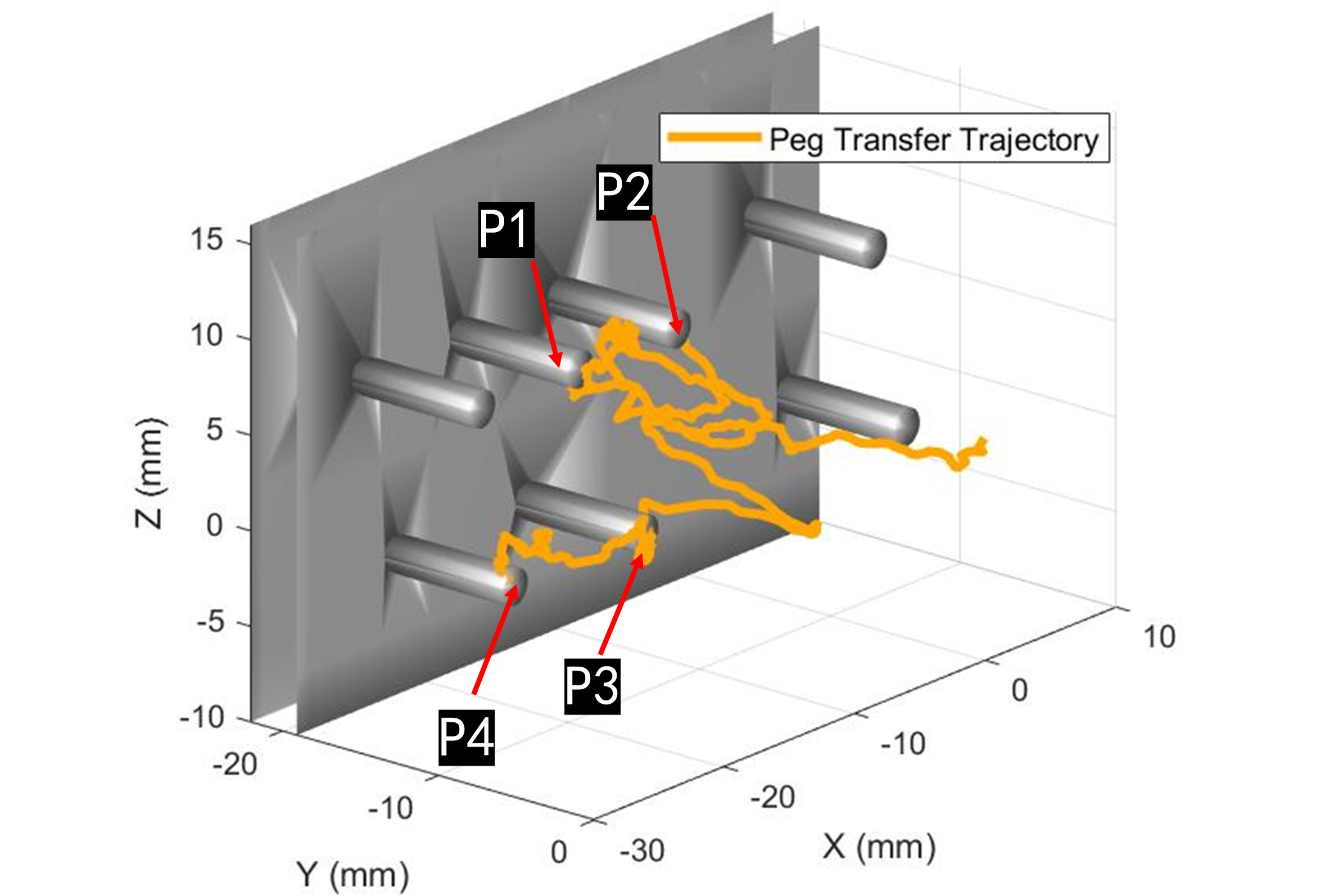}
  \caption{Peg transfer trajectory visualization: the movement of the fastest trajectory process as an example}
  \label{pegTrajectoies}
\end{figure}
% Peg complete process

\begin{figure}[hbt]
  \centering
\includegraphics[width=0.42\textwidth]{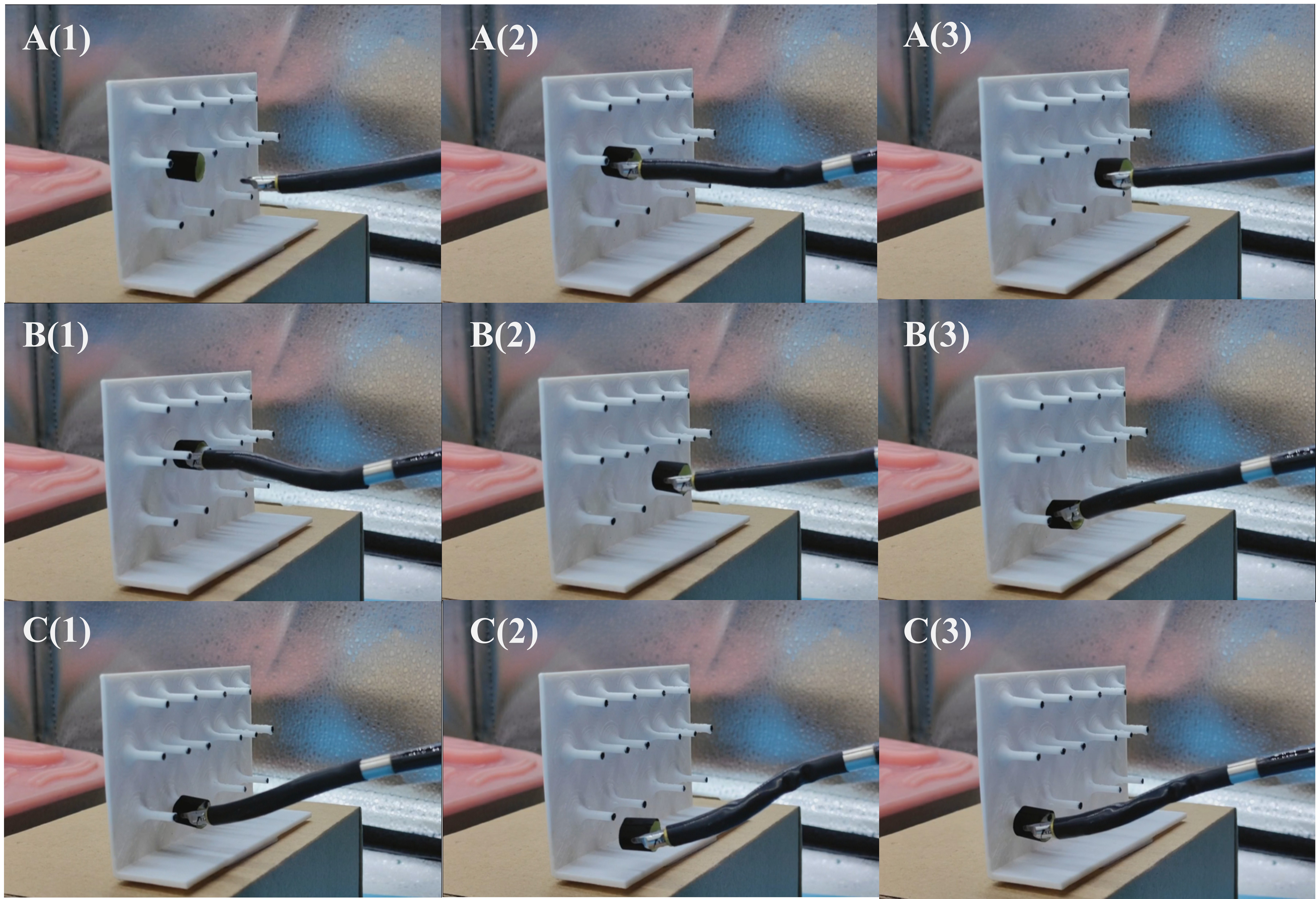}
  \caption{A complete process including four peg transfers: DABM grasps the block at P1 (A(1)→A(2)), transfers and releases it at P2 (A(2)→B(1)), then moves and places it at P3 (B(2)→B(3)), and finally conveys it to P4 (C(1)→C(3))}
  \label{PegTransfer}
\end{figure}
% \vspace{-0.05in}

%% Peg transfer table
\begin{table*}[ht]
  \centering
  \caption{Unilateral Peg‐Transfer Task Experiments}
  \label{tab:peg_transfer_experiments}
  \scalebox{0.90}{% 整体缩放到原来的 90%
  \renewcommand{\arraystretch}{0.8}
    \begin{tabularx}{1.0\linewidth}{
      >{\raggedright\arraybackslash}X
      >{\centering\arraybackslash}X
      >{\centering\arraybackslash}X
      >{\centering\arraybackslash}X
      >{\centering\arraybackslash}X
      >{\centering\arraybackslash}X
    }
      \toprule
      Type & Operator & \mbox{Mean Transfer Time (s)} & Success / Attempts & \mbox{STD for Per Transfer (s)} & Success Rate (\%) \\
      \midrule
      dVRK\cite{hwang2022automating} & 9 volunteers & 12.3 & 610/636 & 12.90 & 95.9  \\
      dVRK\cite{hwang2022automating} & 1 surgeon & \textbf{4.7} & 120/120 & \textbf{1.73} & \textbf{100.0}  \\
      IDCM\cite{wu2019design} & 10 volunteers & 15 & 10/10 & N/A &\textbf{100.0}  \\
      \mbox{GRU-driven DESectBot} & 6 volunteers & 11.8  & 120/120 & 3.93 & \textbf{100.0} \\
      \bottomrule
    \end{tabularx}
  }
\end{table*}

A rectangular base measuring $80$ mm \(\times\) $40$ mm was fabricated with three parallel rows of pegs, each peg $2.25$ mm in diameter. Pegs are spaced $10$ mm apart along the longitudinal (\(Z\)) axis, while the transverse (\(X\)) spacing alternates between $7.5$ mm and $10$ mm. 
Each trial comprised four sequential transfers. Six volunteer participants with no previous surgical experience were recruited; each completed a 15-minute familiarization session prior to formal testing. After system calibration and commissioning, DESectBot was evaluated in a unilateral peg‑transfer task. Figure \ref{pegTrajectoies} depicts the fastest trajectory recorded, and Figure \ref{PegTransfer} illustrates a representative trial:

\begin{enumerate}
  \item \textbf{Pick 1}: From A(1) to A(2), the robot approaches peg~P\(_1\) and grasps the block with its forceps.
  \item \textbf{Transfer 1 / Place 1}: From A(2) to B(1), the block is conveyed to peg~P\(_2\) and released.
  \item \textbf{Pick 2 / Transfer 2 / Place 2}: From B(2) to B(3), the block is retrieved from P\(_2\) and deposited on peg~P\(_3\).
  \item \textbf{Pick 3 / Transfer 3 / Place 3}: From C(1) to C(3), the block is lifted from P\(_3\) and placed on the final peg~P\(_4\).
\end{enumerate}
% Completion of these four transfers constitutes one full peg‑transfer cycle for subsequent performance analysis. Moreover, some potential error need to be discussed first: (1) the systematic error existing in the GRU algorithm; (2) peg placement tolerances and hardware NDI sensor \fcolorbox{gray}{yellow}{R2.3}\hl{itself}; (3) the latency between master hand and DESectBot. For (1), the human operator are subject to the same GRU-driven control pipeline, method systematic error averages out for 120 transfers, so relative success rate and timing comparisons remain valid. For (2), all volunteers operate on the identical physical hardware, the DABM repeatable test ensures consistent clearances for every transfer. For (3), the mean communication latency is less than 40ms, which is much shorter than the given time of 200ms (5hz) for a single motion execution, so it is not affected.

%% 7.24 modified paragragh
Each four‑move sequence constitutes one complete peg‑transfer cycle for analysis, and three potential error sources were considered: (1) any systematic bias in the GRU controller should average out over the 120 transfers performed by each participant, preserving the validity of relative success‑rate and timing comparisons. Second; (2) because every volunteer operates on the same hardware—and repeatability tests confirm consistent peg clearances—the influence of peg‑placement tolerances and the accuracy of the NDI sensor itself is minimal; (3) the mean communication delay between the master device and DESectBot is under $40$ ms, which is significantly lower than the $200$ ms ($5$ Hz) control interval used during evaluation. This confirms that the reported control frequency is not limited by communication latency, but rather by the EM tracking system employed exclusively for trajectory recording and ground-truth comparison.

For benchmarking, the well‑established da Vinci Research Kit (dVRK) and the inverted dual continuum mechanism (IDCM) robot \cite{hwang2022automating,wu2019design} were selected as reference systems. From\cite{hwang2022automating}, the mean transfer time is defined as the overall completion time divided by the number of blocks. Furthermore, the STD for per transfer time is 3.93s, which significantly outperforms the novice-controlled
dVRK systems. The GRU‑driven DESectBot completed the peg‑transfer task faster than volunteers using the dVRK ($11.8$ s vs $12.3$ s) while raising the success rate from $95.9$\% to a perfect 100\%. Relative to volunteers on the IDCM platform, its $11.8$ s mean time is $21.3$\%  faster ($15$ s vs $11.8$ s) with the same flawless success (Table \ref{tab:peg_transfer_experiments}), demonstrating both superior efficiency and reliability without requiring expert operators. 

The combination of sub‑12‑second completion times and zero failures indicates that the GRU‑driven DESectBot delivers reliable and reproducible performance in the peg‑transfer task, supporting its suitability as a standardised platform for ESD skill training.

\begin{figure}[hbt]
  \centering
\includegraphics[width=0.40\textwidth]{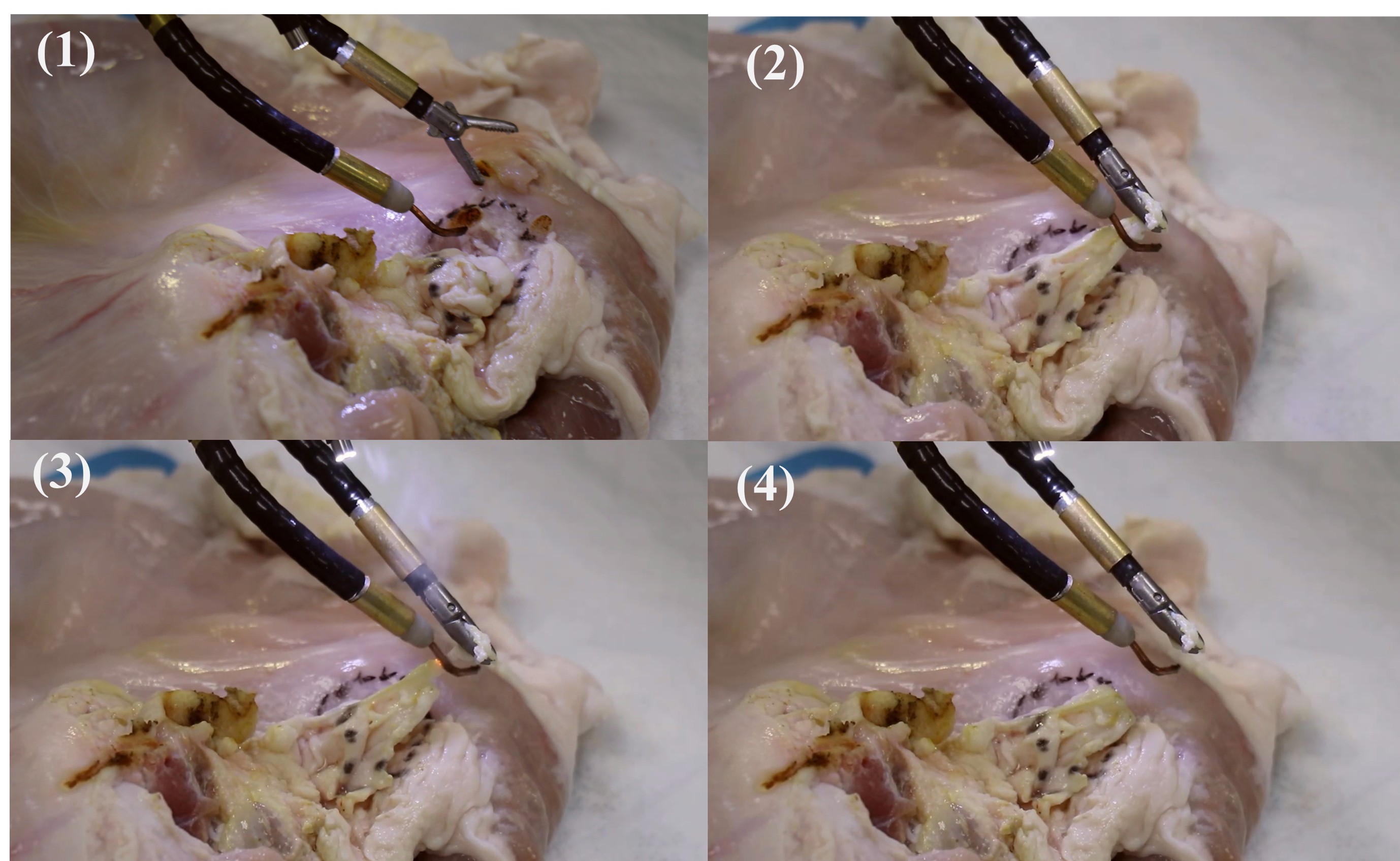}
  \caption{\textit{Ex vivo} evaluation of DESectBot during a simplified ESD procedure on porcine gastric mucosa: (1) positioning; (2) grasping and lifting; (3)→(4) cutting}
  \label{stripMucosal}
  \vspace{-0.1in}
\end{figure}

%% 用forceps剥离表皮组织的实验（主要就是demonstration流程做一下，拍一些照片记录），不需要死磕指标（更多是一个demo的形式），展示其和surgical的relevance即可
% \hl{Furthermore, to validate the performance of DESectBot, an ex vivo experiment was carried out to demonstrate the stripping of porcine mucosa tissue of stomach. One point needs to be mentioned here: the actuator structure and control algorithm of the electrocoagulation scalpel are different from those of forceps. Here, a simplified version of the ESD stripping process is supplemented and presented, the porcine mucosa has been pre-removed in large parts along the marked edge with an electrocoagulation scalpel. From Fig.\ref{stripMucosal} (1) to (4), it shows the process of DESectbot using forceps in conjunction with an electrocoagulation scalpel to grasp and lift tissue and ultimately strip and cut it. This successful ESD steps show that the DESectBot has enough stiffness to cut a thick gastric mucosa and sufficient workspace to cover a big lesion.}
Furthermore, to assess DESectBot’s manipulation capabilities under real clinical settings, an \textit{ex vivo} trial, in which porcine gastric mucosa was dissected, was conducted. 
In light of the differing actuator and control mechanisms between the electro-coagulation scalpel and the forceps, a simplified ESD scenario was established, wherein most of the mucosa was pre-excised along the designated margin using the scalpel. The overall cutting duration is 403 s.
%Because the electro‑coagulation scalpel employs a different actuator and control scheme from the forceps, a simplified ESD scenario was devised: most of the mucosa was pre‑excised along the marked border with the scalpel.
Figures \ref{stripMucosal} (1)-(4) illustrate the grasping, elevating, and finally resecting of the tissue by DESectBot as the scalpel completes the cut, which are the key steps in the ESD procedure. The successful sequence confirms that the robot provides sufficient stiffness to divide thick gastric mucosa and an adequate workspace to address lesions.

\vspace{-0.1in}

\section{Conclusion And Future Work}

% In this paper, a novel dual‐segment continuum robot with an integrated surgical forceps end effector (5 mm diameter, 20 mm length) called DESectBot was established. However, the dual-segment
% structure introduces increased nonlinearities and hysteresis, and the need for more motors to control each segment adds complexity to maintain accuracy. 

In this paper, a novel dual‐segment continuum robot with an integrated surgical forceps end effector named DESectBot was established. For simultaneously control the DABM tip position and orientation, two model-based and three data-driven methods including the proposed GRU were established. The GRU controller consistently outperformed with Jacobian, MPC, FNN, and LSTM method in all trajectory tracking experiments. On a nested‐rectangle path, GRU attained mean positional RMSE $\leq 0.17$ mm and orientational RMSE $\leq 0.96$°, representing improvements of all over the methods. For a complex Lissajous trajectory, GRU achieved 0.81 mm positional and 2.59° orientational RMSE, also surpassing all other methods and demonstrating its ability to compensate for dual-segment continuum nonlinearities. Moreover, in the orientation control with four fixed tip position experiment, the GRU controller markedly outperforms both model-based and alternative data-driven approaches in terms of positional and orientational accuracy and repeatability. 

Building on this, a GRU‑driven peg‑transfer benchmark yielded a $100$\% success rate (120/120) with an average transfer time of $11.8$ s and STD for per transfer time 3.93s, outperforming novice use of the dVRK and matching state‑of‑the‑art continuum systems (IDCM: $15$ s, $100$\% success), supporting suitability for standardized ESD skills training. Furthermore, an \textit{ex vivo} demonstration on porcine gastric mucosa—grasping, elevating, and resecting the tissue while the scalpel completed the cut (total $403$ s)—confirmed that DESectBot provides sufficient stiffness to divide thick mucosa and an operative workspace adequate to address lesions. This cutting duration is comparable to reported values in robotic ESD systems (e.g., $570$ s in \cite{gao2024transendoscopic}) and lies well within the range of clinically reported ESD procedure times, which typically span several tens of minutes depending on lesion size and complexity \cite{ko2022clinical}.

Current limitations include the control update rate used during experimental evaluation, which was constrained by the EM tracking system integrated for ground-truth measurement rather than by the control or communication pipeline itself. In future work, higher control frequencies will be pursued through the integration of faster pose-sensing modalities (e.g., next-generation EM or alternative sensing technologies) and further optimization of the control and communication pipeline, including inference acceleration and real-time software integration. In parallel, the integration of haptic and shape sensing for enhanced feedback, real-time self-calibration to compensate for tendon hysteresis and encoder drift, and the recruitment of professional surgeons for standardized comparative studies are foreseen. In addition, explicit workspace coordination and collision avoidance between the continuum robot and the endoscope will be investigated to mitigate potential workspace conflicts in multi-instrument ESD scenarios. Additional intelligent assistance through image-based lesion detection and motion scaling, together with comprehensive preclinical animal trials, will further advance DESectBot toward robust clinical translation.

% Therefore, in view of the proposed GRU controller has shown superior kinematic registration and control precision in both position and orientation control. The peg transfer test was implemented by the GRU-driven DESectBot, novice users operating DESectBot under GRU control achieved a 100\% success rate (120 success/120 attempts), with an average transfer time of 11.8s which is faster and more reliable than the da Vinci Research Kit (dVRK) under equivalent novice conditions (88.3\% success, 12.3 s) and also comparable to expert performance. The system also matched state‐of‐the‐art continuum benchmarks (IDCM, 15s, 100\% success), confirming its suitability for standardized ESD skill training.\par

In the near future, the integration of haptic and shape sensors for enhanced feedback for DESectBot and real‐time self‐calibration to compensate for tendon hysteresis and encoder drift will be focused on. In parallel, the recruitment of additional professional surgeons is foreseen to enable more standardized comparative assessments. Moreover, intelligent assistance through image‐based lesion detection and motion scaling will also be explored and undertake comprehensive preclinical animal trials to validate performance under realistic ESD conditions.  
These developments will advance DESectBot toward robust clinical translation and broaden its applicability.

\vspace{-0.1in}

\bibliographystyle{IEEEtran}
\bibliography{references}
\vspace{-0.5in}
\begin{IEEEbiography}
[{\includegraphics[width=1in,height=1.25in,clip,keepaspectratio]{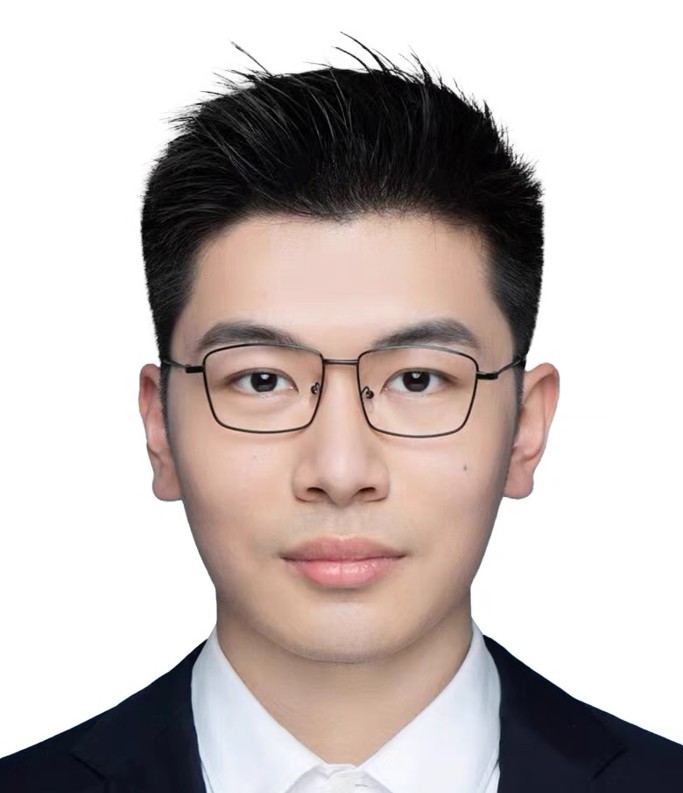}}]
{Yuancheng Shao} (Student Member, IEEE) received the M.Sc. degree in human and biological robotics from Imperial College London in 2021. From 2022 to 2023, he was a medical robot motion planning and control algorithm engineer at the Advanced Technology Institute of Suzhou. He is currently working toward a Ph.D. degree at the City University of Macau in the SIAT-CityU Macau Joint Laboratory. His current research interests include medical robotics and surgical automation techniques.
\end{IEEEbiography}
\vspace{-0.5in}

\begin{IEEEbiography}
[{\includegraphics[width=1in,height=1.25in,clip,keepaspectratio]{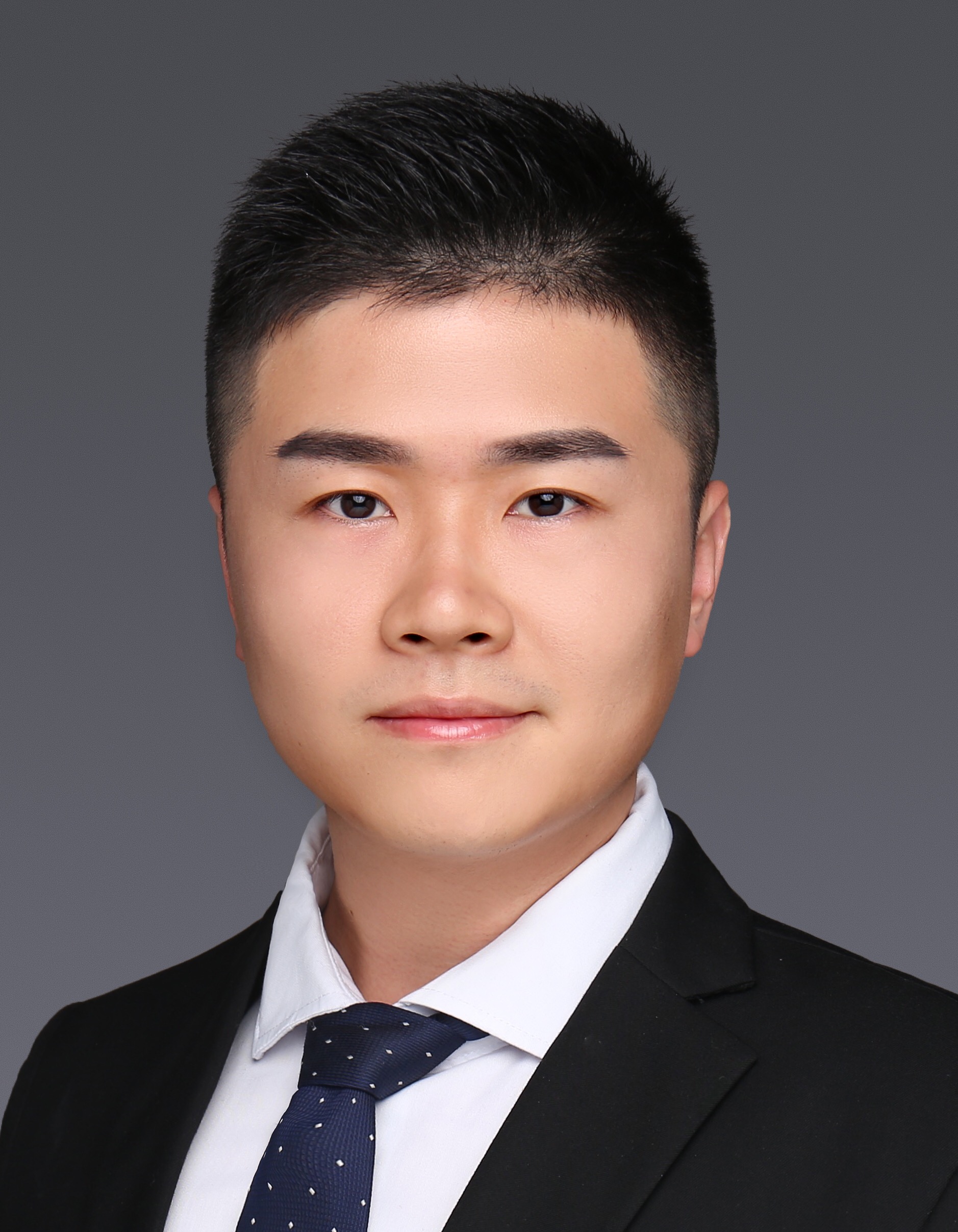}}]
{Yao Zhang} received the M.Sc. degree in medical technology and engineering from the Technical University of Munich, Munich, Germany in 2021. He is currently working toward the Ph.D. degree with Katholieke Universiteit Leuven, Leuven, Belgium. His research interests include novel sensing, robot control, and machine learning in medical robotics.
\end{IEEEbiography}

\vspace{-0.4in}
\begin{IEEEbiography}
[{\includegraphics[width=1in,height=1.25in,clip,keepaspectratio]{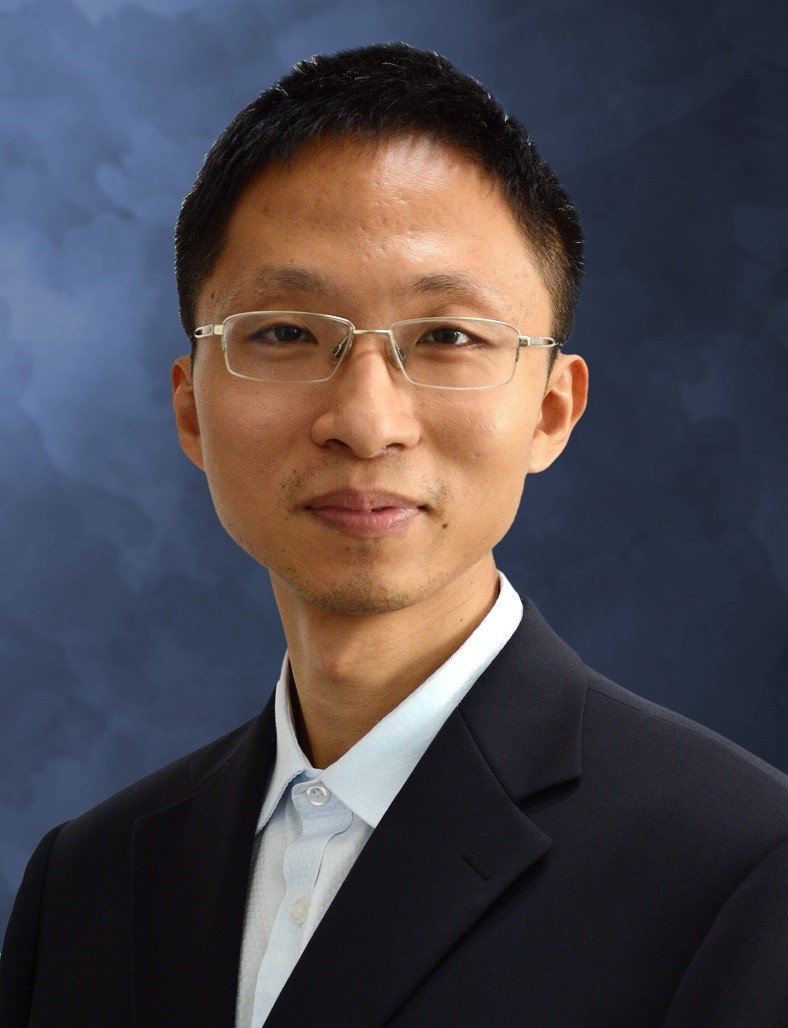}}]
{Jia Gu} (Senior Member, IEEE) is now working with City University of Macau as a full professor and Ph.D. advisor. He received the Ph.D. degree in 2005 from University of Rennes (France). Since 2008, he has joined Chinese Academy of Sciences as a full professor, and then the deputy director of Advanced Technology Institute of Suzhou. His research interests focus on Image-Guided Instruments and Medical Robotics.
\end{IEEEbiography}
\vspace{-0.4in}

\begin{IEEEbiography}
[{\includegraphics[width=1in,height=1.25in,clip,keepaspectratio]{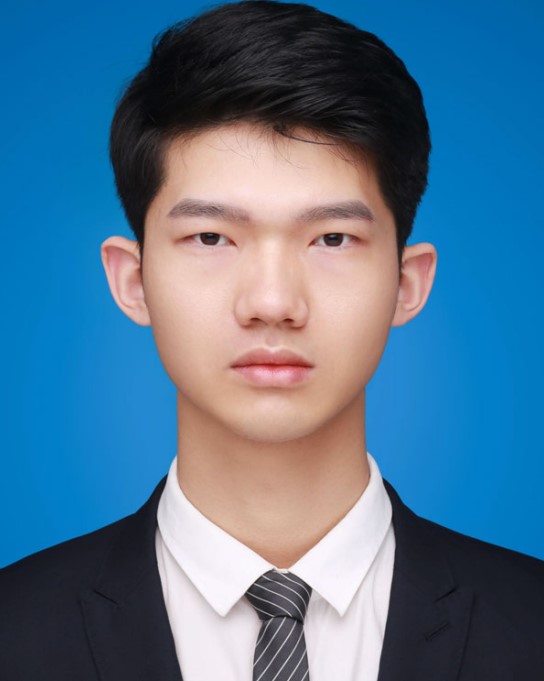}}]
{Zixi Chen} (Graduate Student Member, IEEE) received the M.Sc. degree in control systems from Imperial College in 2021. He is currently pursuing the Ph.D. degree in biorobotics from Scuola Superiore Sant’Anna of Pisa. His research interest includes optical tactile sensors and soft robot control with neural networks.
\end{IEEEbiography}
\vspace{-0.3in}

\begin{IEEEbiography}
[{\includegraphics[width=1in,height=1.25in,clip,keepaspectratio]{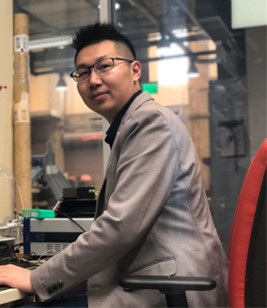}}]
% the M.Sc. degree in medical technology and engineering, from the Technical University of Munich (TUM), Germany, in 2019, and 
{Di Wu} received the PhD degree in 2023 through a joint program between KU Leuven, Belgium, and Delft University of Technology, Netherlands, under the Marie Skáodowska-Curie Actions of the European Union Horizon 2020. He is currently an Assistant Professor at the Maersk Mc-Kinney Møller Institute, University of Southern Denmark. His research interests include surgical robotics, robot control, and machine learning.
\end{IEEEbiography}
\vspace{-0.3in}

\begin{IEEEbiography}
[{\includegraphics[width=1in,height=1.25in,clip,keepaspectratio]{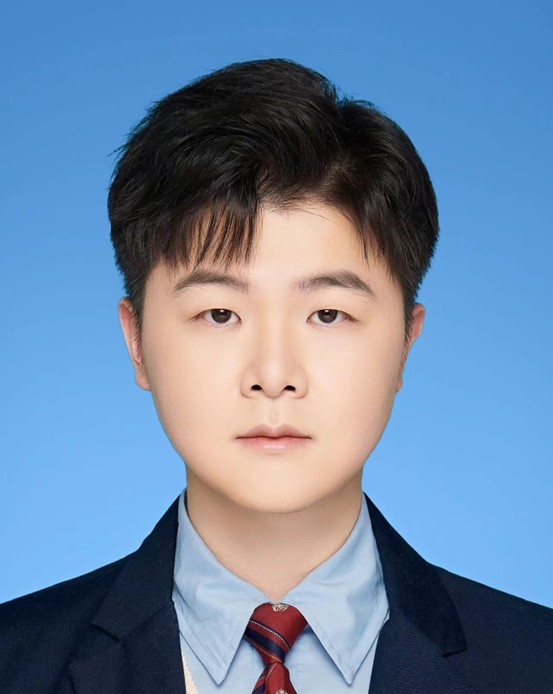}}]
{Yuqiao Chen} received the MEngSc degree in Biomedical Engineering from the University of New South Wales, Sydney, Australia, in 2018. He is currently pursuing a Ph.D. degree in Data Science at the City University of Macau. His research focuses on artificial intelligence and image pattern recognition algorithms in the field of medical robotics.
\end{IEEEbiography}
\vspace{-0.3in}

\begin{IEEEbiography}
[{\includegraphics[width=1in,height=1.25in,clip,keepaspectratio]{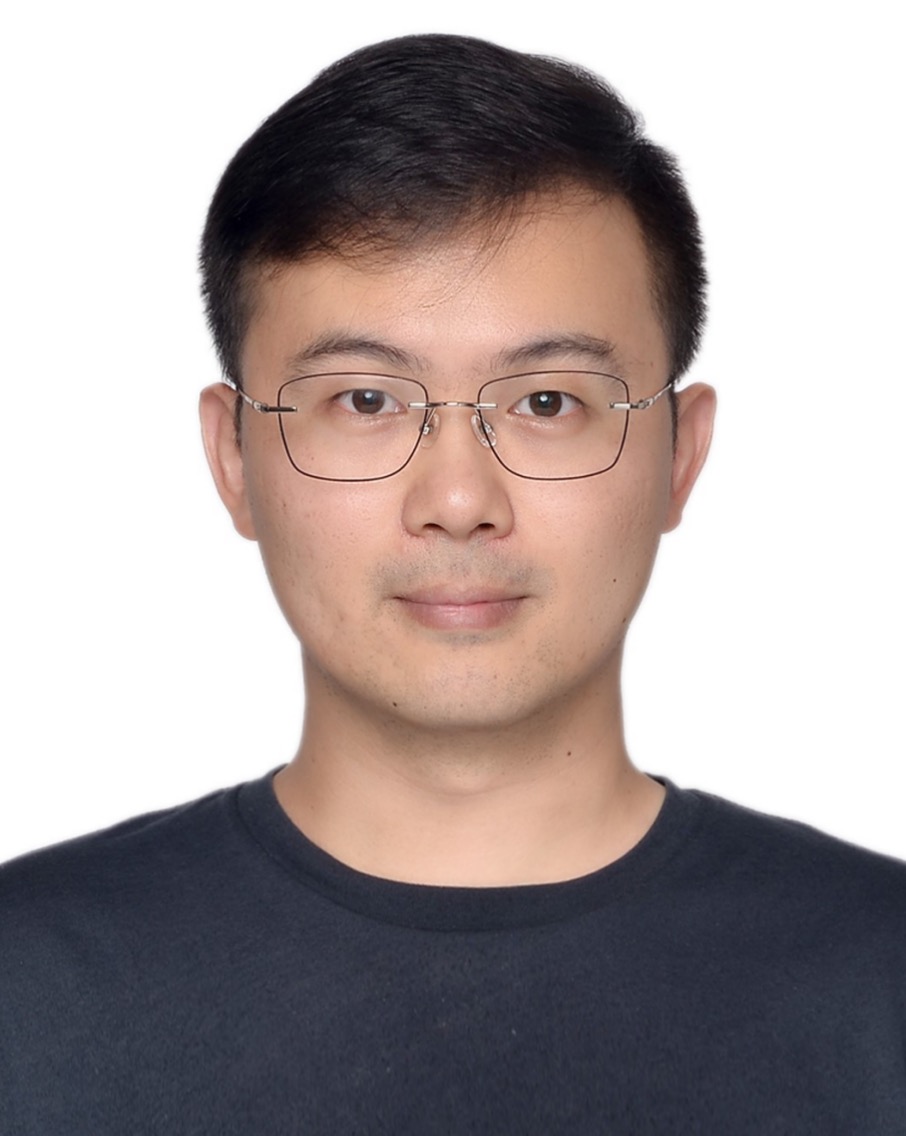}}]
{Bo Lu} (Member, lEEE) received the PhD degree in mechanical engineering from The Hong Kong Polytechnic University, in 2019.  He is currently an Associate Professor with the School of Mechanical and Electrical Engineering, Robotics and Microsystems Center, Soochow University, China. His current research interests include medical robotics, computer vision, and surgical automation techniques.
\end{IEEEbiography}
\vspace{-0.3in}

\begin{IEEEbiography}
[{\includegraphics[width=1in,height=1.25in,clip,keepaspectratio]{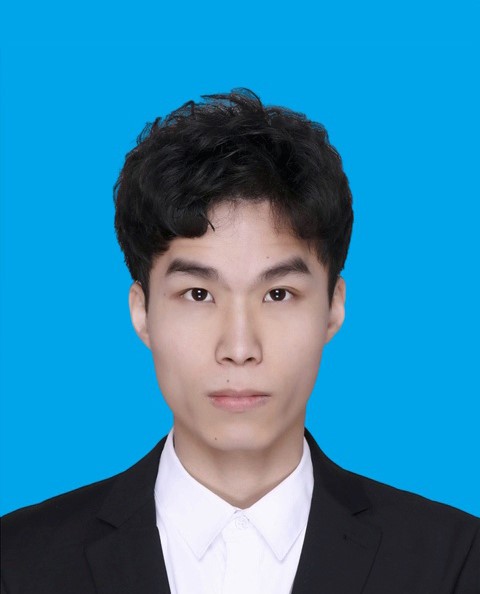}}]
{Wenjie Liu} received the Master degree in the School of Electronic and Information Engineering from Tongji University, China in 2024. His current research interests include medical robotics, and continuum manipulator control.
\end{IEEEbiography}

\vspace{-0.3in} 
\begin{IEEEbiography}
[{\includegraphics[width=1in,height=1.25in,clip,keepaspectratio]{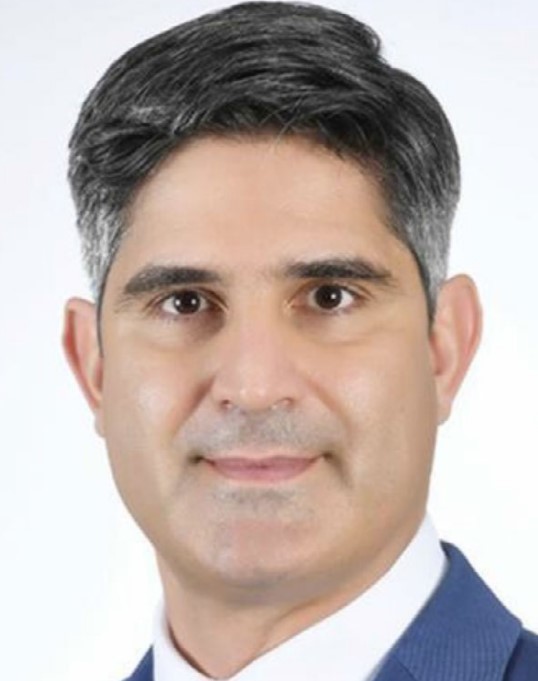}}]
{Cesare Stefanini} (Member, IEEE) received the M.Sc. degree (Hons.) in mechanical engineering and the Ph.D. degree (Hons.) in microengineering from Scuola Superiore Sant’Anna (SSSA), Pisa, Italy, in 1997 and 2002, respectively.  He is currently a professor and the Director of  BioRobotics Institute, SSSA. His research activity is applied to different fields, including underwater robotics, bioinspired systems, biomechatronics, and micromechatronics for medical applications. 
\end{IEEEbiography}

\vspace{-0.3in} 
\begin{IEEEbiography}
[{\includegraphics[width=1in,height=1.25in,clip,keepaspectratio]{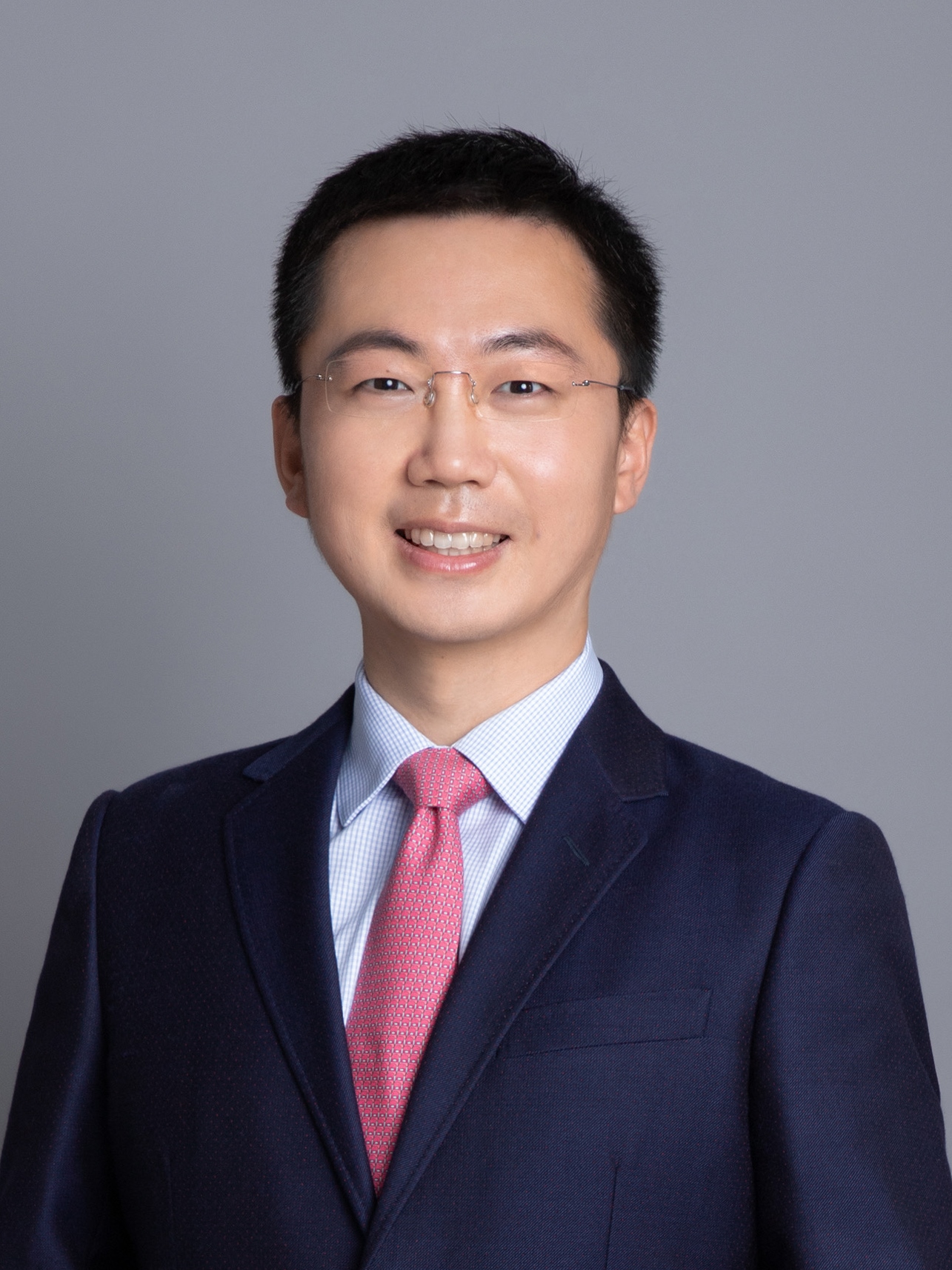}}]
{Peng Qi} (Member, IEEE) received the PhD degree in robotics from King’s College London, U.K., in February 2016. He was a research fellow with the National University of Singapore, from 2015 to 2016, and a visiting scholar (honored) with The Chinese University of Hong Kong, from September 2016 to February 2017. He is currently an associate professor of robotics with Tongji University. His research interests include medical robotics, continuum manipulator, intelligent sensing and interaction.
\end{IEEEbiography}

% \section*{Acknowledgments}


\vfill

\end{document}